\PassOptionsToPackage{table}{xcolor}
\documentclass[letterpaper]{article} 
\usepackage[preprint]{aaai2027} 
\usepackage[hyphens]{url}  
\usepackage{graphicx} 
\graphicspath{{./}{../}}
\usepackage{natbib}  
\usepackage{caption} 
\usepackage{algorithm}
\usepackage{algorithmic}
\usepackage{booktabs}
\usepackage{multirow}
\usepackage{array}
\usepackage{amsmath}
\usepackage{amssymb}
\usepackage{amsfonts}
\usepackage{amsthm}
\newtheorem{proposition}{Proposition}

\definecolor{mygreen}{HTML}{89A090}
\definecolor{pacerblue}{HTML}{D7E7F7}

\definecolor{scorebest}{RGB}{255,243,191}
\definecolor{scorebaseline}{RGB}{223,237,207}
\newcommand{\bestscore}[1]{\begingroup\setlength{\fboxsep}{1pt}\colorbox{scorebest}{\strut#1}\endgroup}
\newcommand{\baselinescore}[1]{\begingroup\setlength{\fboxsep}{1pt}\colorbox{scorebaseline}{\strut#1}\endgroup}
\newcommand{\scoregain}[3]{\mbox{\textbf{#1}{\boldmath$^{\mathord{#2}\!\text{\bfseries#3}}$}}}

\usepackage{newfloat}
\usepackage{listings}
\DeclareCaptionStyle{ruled}{labelfont=normalfont,labelsep=colon,strut=off}  
\lstdefinestyle{jsonnormal}{
  basicstyle=\normalfont\small,
  columns=fullflexible,
  keepspaces=true,
  breaklines=true,
  frame=none,
  showstringspaces=false
}
\floatstyle{ruled}
\newfloat{listing}{tb}{lst}{}
\floatname{listing}{Listing}
\title{Case as Judge: A Policy-Decoupled Framework for Training-Free Content Moderation}
\author{
    Qi Lu\textsuperscript{\rm 1,2},
    Sen Li\textsuperscript{\rm 1},
    Likun Shi\textsuperscript{\rm 1},
    Xiaolong Jiang\textsuperscript{\rm 1}\corresponding
}
\affiliations{
    \textsuperscript{\rm 1}Xiaohongshu\\
    \textsuperscript{\rm 2}University of Science and Technology of China\\
    \{luqi4, lisen2, shilikun, laige\}@xiaohongshu.com
}

\begin{document}

\maketitle

\begin{abstract}

    Content moderation is inherently policy-dependent: moderation boundaries are often subtle, subjective, and frequently updated, while conventional rule based or fine-tuning based systems require costly rule rewriting, model retraining, or system redeployment. We propose CaseLaw, a policy-decoupled and training-free framework for adaptive content moderation. CaseLaw reformulates moderation as precedent-based retrieval and voting, where multimodal content is transformed into evidence-preserved structured semantics, embedded into a stable semantic space, and classified by thresholded voting over retrieved precedent labels. We train NoteCap on a large-scale corpus of structurally annotated notes to ensure consistent semantic extraction and freeze it during subsequent policy adaptation, which requires no additional training. Policy adaptation is achieved by updating the semantic case library, including case labels and mappings, without retraining models or modifying prompts. To evaluate CaseLaw, we construct MFO-RedNotes, a benchmark for the male-female opposition risk domain. We additionally provide a five category CoPro comparison. Experiments show that CaseLaw consistently outperforms competing baselines in both precision and recall across policy-dependent risk levels and broader safety categories, with category-wise precision/recall reaching up to 86.8\%/90.1\% on MFO-RedNotes across both policy settings and 94.5\%/92.3\% on CoPro. Our work highlights the potential of precedent-based reasoning for policy adaptive content moderation and offers a practical direction toward more stable, transparent, and training-free moderation systems. \textcolor{red}{MFO-Datasets is available at \url{https://huggingface.co/datasets/luqi4/MFO-Datasets}.}
\end{abstract}

\section{Introduction}

Content moderation has become a foundational mechanism for maintaining safe, inclusive, and trustworthy online ecosystems. Modern platforms must continuously regulate user-generated content at massive scale while balancing platform policies, community norms, legal requirements, and user expectations \cite{gillespie2018custodians,gerrard2020behind,gorwa2020algorithmic}. This challenge has become more urgent with the rapid deployment of large language models (LLMs) and generative AI systems, whose outputs may involve a broad spectrum of harms, including harassment, hate speech, misinformation, sexual content, self-harm, privacy violation, and other emerging sociotechnical risks \cite{shelby2023sociotechnical,markov2023holistic,brown2020language,ouyang2022training,inan2023llama,huang2025content}. Unlike conventional classification tasks, content moderation is inherently policy-dependent: the same content may be judged differently across platforms, cultural contexts, user groups, or temporal policy updates. Therefore, a practical moderation system must not only detect harmful content accurately, but also adapt rapidly when the definition, severity, or boundary of risk changes.

\begin{figure}[t]
    \centering
    \includegraphics[width=\columnwidth]{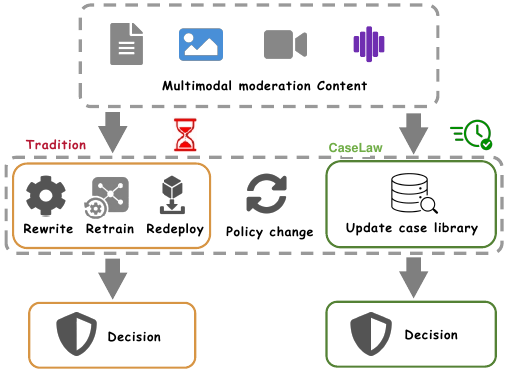}
    \caption{Comparison of policy adaptation in conventional moderation pipelines and CaseLaw. Policy changes in conventional pipelines can require rule rewriting, model retraining, and redeployment of updated moderation components. CaseLaw adapts by updating the external case library while keeping its deployed models fixed.}
    \label{fig:contrast}
\end{figure}

Current automated moderation pipelines largely follow two decision pathways: rule-guided LLM agents and supervised classification models. Rule-guided agents interpret policies in context, extending their coverage beyond explicitly enumerated patterns \cite{palla2025policy}. Their generative decision process, however, can produce inconsistent judgments, particularly for borderline cases: repeated evaluations of identical content with an identical prompt may yield different outcomes. Conventional classifiers operationalize moderation policies through annotated datasets and supervised learning \cite{nobata2016abusive,markov2023holistic}. With fixed parameters and deterministic inference, these models provide repeatable labels and scores, but their learned decision boundaries do not readily accommodate policy revisions. Adapting them often requires retraining or rewriting policy prompt, and redeployment, resulting in long iteration cycles, limited flexibility, and substantial maintenance costs.

Figure~\ref{fig:contrast} illustrates the limitations of conventional moderation pipelines, in which policy revisions necessitate rule rewriting, model retraining, and redeployment. Across both decision pathways, moderation quality also depends on the representation of policy-relevant evidence. LLM-based decision stages are sensitive to wording, formatting, input length, and the position of relevant information within the context \cite{sclar2024quantifying,liu2024lost}, while repeatable classifier outputs likewise do not guarantee robustness to alternative expressions of the same underlying evidence. Although retrieval-augmented methods supply additional context through similar examples \cite{chen2024class}, similarity between raw examples may be insufficient to support fine-grained, policy-dependent judgments. Such judgments often hinge on structured risk cues, rhetorical strategies, and subtle distinctions near policy boundaries that broad semantic resemblance fails to capture.

To address these challenges, we propose CaseLaw, a case-driven, policy-decoupled framework for training-free moderation adaptation. CaseLaw is motivated by the observation that the semantics of content under moderation remain relatively stable, whereas moderation policies, risk boundaries, and enforcement priorities evolve. CaseLaw formulates moderation as case-based semantic inference: it converts heterogeneous user generated content into structured semantic representations, retrieves relevant precedents in a fixed embedding space, and aggregates their policy-dependent labels. Policy changes are expressed as revisions to an external semantic case library, including updates to case labels, case-specific thresholds, and policy-to-label mappings. By moving adaptation from model optimization to explicit case-level knowledge, CaseLaw enables lightweight and controllable policy updates while keeping the representation models fixed. To improve the consistency and fidelity of structured semantic extraction, we annotate 100,000 notes with structured semantic fields and train NoteCap, a dedicated extraction model. NoteCap remains frozen and is reused across policy updates, preserving the semantic evidence required for precedent retrieval while moderation behavior adapts through case-library updates without model training.

We instantiate and evaluate CaseLaw in the gender-polarization risk domain, where violations often depend on nuanced framing, implicit hostility, group opposition, and contextual escalation. We construct MFO-RedNotes, a policy-dependent multimodal moderation benchmark comprising 50,000 notes across four risk levels: Safe, Low-risk, Mid-risk, and High-risk. Of these notes, 10,000 receive different labels following a policy change, enabling evaluation of adaptation to revised decision boundaries while holding the underlying content fixed. Experiments show that CaseLaw consistently outperforms zero-shot prompting, fixed few-shot prompting, fine-tuning, and retrieval-augmented baselines across risk levels under both policy settings, supporting the effectiveness of policy-decoupled, case-based moderation.

Our contributions are summarized as follows:

\begin{itemize}
    \item We propose CaseLaw, a case-driven moderation framework that decouples policy adaptation from model optimization through an editable semantic case library, enabling policy updates without retraining deployed models or rewriting prompts.
    \item We develop an evidence-preserving structured representation and train NoteCap on 100,000 annotated notes to support precedent retrieval and thresholded label voting. The representation models remain fixed throughout all subsequent policy updates.
    \item We construct MFO-RedNotes with 50,000 notes and policy-induced label changes for 10,000 of them, and demonstrate CaseLaw's effectiveness across four risk levels before and after policy change.
\end{itemize}

\begin{figure*}[t]
    \centering
    \includegraphics[width=\textwidth]{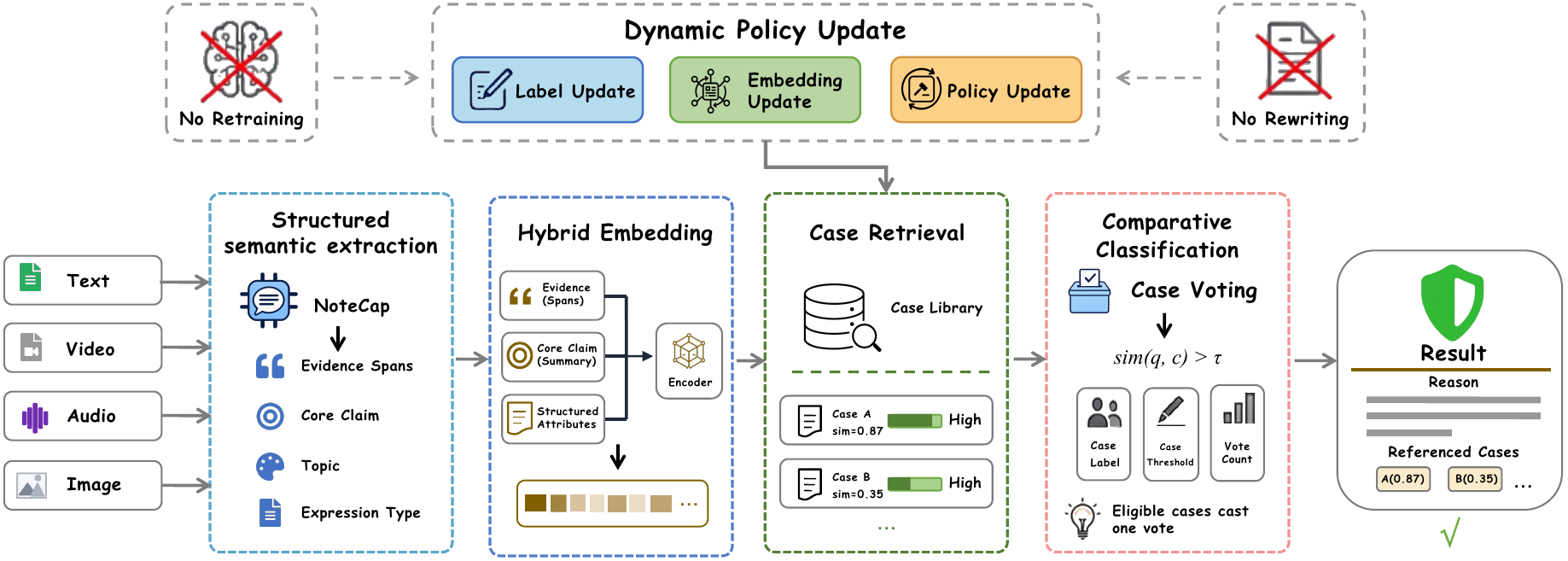}
    \caption{Overview of CaseLaw. NoteCap converts text, images, video, and audio into evidence-preserving structured fields. A frozen encoder embeds the serialized fields, enabling retrieval of precedent cases from an editable library. Each case whose similarity meets or exceeds its stored threshold casts a single vote based on its current risk label, without invoking an LLM classifier after retrieval. Policy changes are implemented by updating precedent labels and metadata while keeping the deployed representation models fixed.}
    \label{fig:caselaw-method}
    \end{figure*}

\section{Related Work}

    \paragraph{Automated content moderation and AI safety.}
    Automated content moderation scales platform governance through keyword filtering, hash matching, and supervised classifiers. Automated approaches remain sensitive to linguistic variation and contextual ambiguity \cite{nobata2016abusive,gorwa2020algorithmic}. Subsequent machine learning approaches addressed harassment, abusive language, hate speech, and offensive content \cite{nobata2016abusive,gamback2017using}. Word embeddings capture lexical semantics \cite{mikolov2013efficient,pennington2014glove}, while Transformer architectures and bidirectional pretraining support contextual representations \cite{vaswani2017attention,devlin2019bert}. However, greater semantic capacity does not eliminate policy dependence: the same content can receive different labels under different governance requirements. Moderation decisions remain shaped by policy taxonomies, community expectations, cultural context, and evolving definitions of harm \cite{palla2025policy,huang2025content}.

    \paragraph{Policy operationalization and prompt-based moderation.}
    Operationalizing abstract policies is central to automated moderation. Traditional pipelines translate policies into annotation guidelines and labeled datasets for supervised training \cite{markov2023holistic,palla2025policy}. LLMs support an alternative approach in which policies are supplied directly through prompts, allowing moderation criteria to change without constructing a new training dataset for every policy revision \cite{brown2020language,ouyang2022training,palla2025policy}. Chain-of-thought prompting and ReAct support multi-step reasoning and the integration of reasoning with actions, respectively \cite{wei2022chain,yao2022react}. However, LLM judgments are sensitive to prompt formatting, input length, and the position of relevant information \cite{sclar2024quantifying,liu2024lost}. This sensitivity complicates consistent enforcement across policy versions: even minor changes in policy wording or presentation can alter moderation decisions for otherwise unchanged content.

    \paragraph{Retrieval-augmented generation and case-based moderation.}
    Retrieval-augmented generation (RAG) augments pretrained language models with external non-parametric memory, enabling models to access updatable knowledge without changing their parameters \cite{lewis2020retrieval}. RAG has been widely studied for knowledge-intensive tasks and has been shown to improve factuality, transparency, and adaptability by retrieving relevant external evidence \cite{gao2023retrieval}. Dense retrieval methods further improve the quality of retrieved evidence through representation learning and contrastive training \cite{lin2023train}, while libraries such as Faiss enable efficient large-scale similarity search over embedding spaces \cite{douze2025faiss}. In content moderation, Class-RAG~\cite{chen2024class} uses a library of safe and unsafe examples to support real-time semantic hotfixing and improve robustness to adversarial perturbations. 

\section{Method}

As illustrated in Figure~\ref{fig:caselaw-method}, CaseLaw transforms multimodal content into evidence-preserving structured representations, embeds them in a retrieval-oriented semantic space, and retrieves similar precedents from an editable case library. Each retrieved precedent casts one vote for its current policy-dependent label only when its similarity to the query meets or exceeds its own stored threshold. The extractor, encoder, extraction prompt, retrieval configuration, and voting rule remain fixed throughout deployment. Policy adaptation proceeds solely through updates to the external case library, allowing moderation behavior to track changing policies without retraining the deployed models.

    \subsection{Problem Formulation}

    Let $x \in \mathcal{X}$ denote a user-generated content instance, which may contain multimodal content such as text, images, videos, and audio. CaseLaw maps $x$ to a moderation classification result defined over a user-specified label space:
    \begin{equation}
    \mathcal{Y}_t=\{\ell^{(t)}_1,\ell^{(t)}_2,\ldots,\ell^{(t)}_{M_t}\},
    \end{equation}
    where each $\ell^{(t)}_j$ denotes a policy-dependent moderation label at moderation time $t$, and $M_t$ is the number of labels under the current policy configuration. The label space is not assumed to be fixed and can be customized for different moderation scenarios, including risk levels, violation categories, governance actions, or domain-specific policy outcomes.
    
    At moderation time $t$, the effective policy state is denoted as $p_t$. A change from $p_t$ to $p_{t+1}$ may introduce new labels, revise the semantic boundary of existing labels, or alter the mapping between content patterns and moderation outcomes. Let $\mathcal{D}_t=\{(x_i,y_i^{(t)})\}_{i=1}^{n_t}$ denote the evaluation distribution under policy state $p_t$, where $y_i^{(t)}\in\mathcal{Y}_t$ is the policy-dependent ground-truth label. The goal of CaseLaw is to maintain accurate moderation decisions across a sequence of policy states without retraining the deployed representation models or modifying the extraction prompt.
    
    Formally, let $f_{\Theta,\pi}$ denote the moderation function, where $\Theta$ represents the parameters of all frozen model components and $\pi$ denotes the fixed prompt template. Given a policy state $p_t$, let $\mathcal{D}_t$ denote the corresponding evaluation distribution, and let $\mathcal{Y}_t$ be the policy-dependent label space at time $t$. For a case library $\mathcal{L}_t$, the expected classification accuracy under policy state $p_t$ is defined as
    \begin{equation}
    \begin{aligned}
    \operatorname{Acc}_t(\mathcal{L}_t)
    &=
    \mathbb{E}_{(x,y^{(t)})\sim \mathcal{D}_t}
    \left[
    \mathbb{I}
    \left\{
    f_{\Theta,\pi}(x;\mathcal{L}_t,\mathcal{Y}_t)=y^{(t)}
    \right\}
    \right],
    \end{aligned}
    \end{equation}
    where fixed $\Theta$ and $\pi$ are suppressed in the notation and $\mathbb{I}\{\cdot\}$ is the indicator function, which equals $1$ if the predicted label matches the policy-dependent ground-truth label and $0$ otherwise.
    
    CaseLaw aims to maintain high classification performance across dynamic policy states by updating only the external case library:
    \begin{equation}
    \max_{\{\mathcal{L}_t\}_{t=1}^{T}}
    \frac{1}{T}
    \sum_{t=1}^{T}
    \operatorname{Acc}_t(\mathcal{L}_t),
    \end{equation}

    subject to the training-free and prompt-invariant constraints:
    $\Theta_{t+1}=\Theta_t$, $\pi_{t+1}=\pi_t$, while
    $\mathcal{Y}_{t+1}$ is allowed to differ from $\mathcal{Y}_t$.

    \subsection{Structured Semantic Extraction}
    
    Content moderation often depends on subtle linguistic cues such as group attribution, sarcasm, metaphor, emotionally charged expressions, and risk-bearing lexical choices. Directly embedding the full raw content may introduce substantial noise, while embedding a purely abstractive LLM summary may remove precisely the words that determine risk severity. CaseLaw therefore introduces an evidence-preserved structured semantic extraction module.
        
    We obtain NoteCap through supervised training on 100,000 notes annotated with the same structured schema used at inference. The supervision targets semantic organization and evidence preservation rather than a particular policy's final risk labels. Once trained, NoteCap is frozen across policy versions. 

    For multimodal inputs, OCR extracts text appearing in images, and ASR transcribes speech in video and audio. These textual observations are combined with the original images and sampled video frames as multimodal inputs to evidence selection. OCR and ASR supplement visual evidence. This preserves contextual cues that may not be recoverable from transcripts alone.

    For each input $x$, the frozen NoteCap extractor produces a structured representation
    \begin{equation}
    a_x = \mathcal{S}_{\theta}(x)
    = \bigl(E_x, C_x, T_x, B_x, M_x\bigr),
    \end{equation}
    where $E_x$ denotes evidence spans, $C_x$ denotes the core claim, $T_x$ denotes the topic, $B_x$ denotes the subtopic, and $M_x$ denotes the expression type. Evidence spans are selected from the original content and are required to preserve risk-bearing phrases, entity references, and potentially controversial claims. In practice, CaseLaw retains a small set of suspicious original spans, typically six to eight sentences or fragments, and combines them with a concise core claim of approximately 150--200 words.
        
    The extraction module is designed to optimize a trade-off between compression and evidence preservation. Let $\mathcal{G}(x)=\{g_j\}_{j=1}^{m}$ be candidate textual units extracted from the multimodal content, including text, images, videos, and audio information. CaseLaw selects evidence spans according to
    \begin{equation}
    E_x=\operatorname{Top}_{m_e}
    \left\{
    g_j \in \mathcal{G}(x):
    \rho_{\theta}(g_j,x) 
    \right\},
    \end{equation}
    where $\rho_{\theta}(g_j,x)$ is the extractor's estimated relevance of candidate span $g_j$ to potential moderation risk, and $m_e$ controls the number of retained evidence spans.
        
    The final retrieval text is constructed by serializing the structured representation $a_x$, which jointly contains verbatim evidence and abstractive semantic components. Specifically, the evidence spans preserve lexical and rhetorical signals, while the core claim and structured attributes provide a compact semantic abstraction. This design differs from standard summarization-based retrieval: rather than relying solely on an abstractive summary, CaseLaw preserves risk-bearing original expressions while maintaining a concise semantic representation. As a result, CaseLaw reduces semantic drift caused by purely abstractive summaries and improves the retrieval of moderation-relevant precedent cases.
    
    \subsection{Case Library Construction}

    The case library serves as the policy-bearing knowledge layer of CaseLaw. Each case stores its original content, structured semantics, embedding representation, moderation label, decision rationale, and similarity threshold. Formally, at moderation time $t$, the case library is denoted as
    \begin{equation}
    \mathcal{L}_t=\{c_i\}_{i=1}^{N_t}, \quad
    c_i=(u_i,a_i,z_i,y_i,r_i,\tau_i),
    \end{equation}
    where $u_i$ is the original case content, $a_i$ is the structured semantic representation, $z_i$ is the retrieval embedding, $y_i\in\mathcal{Y}_t$ is the policy-dependent moderation label, $r_i$ is the decision rationale, and $\tau_i$ is the case-specific similarity threshold for voting eligibility. This formulation allows policy updates to be implemented through case-level revisions, such as label correction, threshold recalibration, or case insertion, without modifying model parameters or rewriting the prompt.

    Case-specific thresholds account for differences in how strongly a case's semantics indicate its associated risk. Cases with implicit risk cues require more selective matching: a higher threshold restricts their votes to closer semantic matches and helps avoid incorrectly flagging safe notes. Before admitting a training-set case, we calibrate its threshold on the validation set, which contains notes labeled as safe or violating under the applicable policy. We assess candidate threshold values according to their effects on violating-note recall and the false-positive rate on safe notes, and select an operating point for each case that balances these two criteria.

    For each case $c_i$, CaseLaw applies the same structured extraction and serialization procedure as used for the query. Since risk-bearing lexical evidence has already been preserved as evidence spans during structured semantic extraction, CaseLaw does not introduce an additional sparse lexical representation. Instead, the serialized representation $\tau(a_i)$ is directly encoded into a dense semantic embedding:
    \begin{equation}
    z_i=\mathcal{E}_{\phi}(\tau(a_i)), \quad z_i\in\mathbb{R}^{D},
    \end{equation}
    where $\mathcal{E}_{\phi}$ is a frozen embedding encoder. Because $\tau(a_i)$ contains both verbatim evidence spans and abstractive semantic fields, the resulting embedding captures not only the overall semantic meaning of the case but also the risk-bearing expressions preserved from the original content.

    Given a query $x$, CaseLaw retrieves similar precedent cases through approximate nearest neighbor (ANN) search \cite{arya1998optimal} over the current case library. Each case embedding is indexed in advance, and the query embedding $z_x$ is used to efficiently retrieve the top-$k$ nearest cases according to embedding similarity. Since the embedding input is constructed from evidence-preserved structured semantics, including evidence spans, topic, subtopic, expression type, and core claim, ANN retrieval is performed over a moderation-oriented semantic space rather than raw content alone. The retrieved precedent set is then obtained by

    \begin{equation}
    \mathcal{N}_k(x)
    =
    \operatorname{Top}_{k}
    \left\{
    \operatorname{Sim}(x,c_i): c_i\in\mathcal{L}_t
    \right\}.
    \end{equation}

    Benefiting from the evidence-preserved structured semantic extraction, the retrieval representation retains both risk-bearing original expressions and compact semantic attributes, enabling semantically similar content to be retrieved more precisely. The retrieved top-$k$ cases then serve as case-level evidence for the downstream voting module, in which their labels determine the prediction, while the stored rationales remain available for auditing without contributing additional votes.
    
    \subsection{Thresholded Precedent Voting}
    \label{sec:voting}

    CaseLaw computes its final decision directly from the current labels of retrieved precedents. For a query embedding $z_x$, let $s_i=\operatorname{cos}(z_x,z_i)$. Each retrieved case $c_i$ participates in voting only if its similarity score is greater than or equal to its own stored threshold $\tau_i$. The eligible voting set is
    \begin{equation}
    \mathcal{V}_{k}(x)=\{c_i\in\mathcal{N}_k(x):s_i\geq\tau_i\}.
    \label{eq:eligible-votes}
    \end{equation}
    Each eligible case contributes one vote to its current label. Let $\mathcal{V}_{\ell}(x)=\{c_i\in\mathcal{V}_{k}(x):y_i^{(t)}=\ell\}$ denote the eligible cases supporting label $\ell$, and let $n_{\ell}=|\mathcal{V}_{\ell}(x)|$ be their vote count. The final label is selected by
    \begin{equation}
    \hat y^{(t)}=
    \operatorname*{arg\,max}_{\ell\in\mathcal{Y}_t:\,n_{\ell}>0}^{\mathrm{lex}}
    \left(n_{\ell},\frac{\sum_{c_i\in\mathcal{V}_{\ell}(x)}s_i}{n_{\ell}}\right),
    \label{eq:vote-count}
    \end{equation}
    where lexicographic maximization prioritizes the vote count and uses the mean similarity of the supporting eligible cases to break ties. If no retrieved case meets its threshold, the instance is referred for human review.

    We mitigate residual label ties in both vote count and mean similarity by adjusting the retrieval depth $k$ and expanding the case library. With $k=16$ and a library of 40,000 cases, we observed no ties among the highest-ranked labels in our evaluation. Any remaining cases tied on both criteria are referred for human review.

    \begin{table*}[t]
        \centering
        \caption{\textbf{Comparison on MFO-RedNotes before and after policy change.}
        Precision and recall are expressed as percentages for four policy-dependent risk levels.
        Yellow highlights the best score, and green highlights the strongest baseline in each column.
        Superscripts report absolute gains in percentage points over the strongest baseline within the same policy setting.}
        \label{tab:risk_level_results}
        \footnotesize
        \setlength{\tabcolsep}{2.5pt}
        \renewcommand{\arraystretch}{1.12}
        \newcommand{\pavmetricwidth}{\dimexpr(\textwidth-.19\textwidth-17\tabcolsep-4\arrayrulewidth)/8\relax}
        \begin{tabular*}{\textwidth}{@{}>{\raggedright\arraybackslash}p{.19\textwidth}|
            *{3}{>{\centering\arraybackslash}p{\pavmetricwidth}>{\centering\arraybackslash}p{\pavmetricwidth}|}
            >{\centering\arraybackslash}p{\pavmetricwidth}>{\centering\arraybackslash}p{\pavmetricwidth}}
            \toprule
            \multirow{2}{*}{\textbf{Method}} & \multicolumn{2}{c|}{\textbf{Safe}} & \multicolumn{2}{c|}{\textbf{Low-risk}} & \multicolumn{2}{c|}{\textbf{Mid-risk}} & \multicolumn{2}{c}{\textbf{High-risk}} \\
            \cmidrule(lr){2-3}\cmidrule(lr){4-5}\cmidrule(lr){6-7}\cmidrule(lr){8-9}
            & Prec (\%) & Rec (\%) & Prec (\%) & Rec (\%) & Prec (\%) & Rec (\%) & Prec (\%) & Rec (\%) \\
            \midrule
            \rowcolor{gray!12}\multicolumn{9}{c}{\textbf{\textit{Original policy}}} \\
            \midrule
            Zero-shot & 56.6 & 54.5 & 39.7 & 41.0 & 52.0 & 53.0 & 50.7 & 50.7 \\
            Fixed few-shot & 71.5 & 72.2 & 55.4 & 56.0 & 64.6 & 63.4 & 68.1 & 67.8 \\
            Fine-tuning & 78.1 & 85.0 & 73.3 & 71.0 & 76.3 & 74.0 & 81.2 & 76.2 \\
            Class-RAG & \baselinescore{81.0} & \baselinescore{85.3} & \baselinescore{75.8} & \baselinescore{74.5} & \baselinescore{78.5} & \baselinescore{76.3} & \baselinescore{82.5} & \baselinescore{78.3} \\
            Llama Guard & 66.3 & 65.7 & 49.6 & 52.3 & 58.8 & 56.7 & 60.3 & 60.3 \\
            \midrule
            \rowcolor{gray!12}\textbf{CaseLaw (ours)} & \bestscore{\scoregain{85.3}{\uparrow}{4.3}} & \bestscore{\scoregain{88.3}{\uparrow}{3.0}} & \bestscore{\scoregain{82.9}{\uparrow}{7.1}} & \bestscore{\scoregain{76.0}{\uparrow}{1.5}} & \bestscore{\scoregain{84.8}{\uparrow}{6.3}} & \bestscore{\scoregain{81.0}{\uparrow}{4.7}} & \bestscore{\scoregain{85.3}{\uparrow}{2.8}} & \bestscore{\scoregain{90.1}{\uparrow}{11.8}} \\
            \midrule
            \rowcolor{gray!12}\multicolumn{9}{c}{\textbf{\textit{After policy change}}} \\
            \midrule
            Zero-shot & 57.2 & 55.3 & 41.3 & 39.4 & 39.7 & 43.5 & 45.4 & 46.2 \\
            Fixed few-shot & 64.2 & 62.1 & 47.9 & 53.4 & 47.6 & 47.0 & 56.6 & 52.5 \\
            Fine-tuning & 66.7 & 67.5 & \baselinescore{65.2} & \baselinescore{63.4} & 61.1 & 65.4 & 67.5 & \baselinescore{66.2} \\
            Class-RAG & \baselinescore{68.2} & \baselinescore{69.3} & 65.1 & 62.7 & \baselinescore{62.3} & \baselinescore{66.7} & \baselinescore{68.3} & 65.8 \\
            Llama Guard & 61.0 & 59.5 & 44.6 & 49.2 & 43.7 & 42.3 & 51.9 & 49.3 \\
            \midrule
            \rowcolor{gray!12}\textbf{CaseLaw (ours)} & \bestscore{\scoregain{86.8}{\uparrow}{18.6}} & \bestscore{\scoregain{88.0}{\uparrow}{18.7}} & \bestscore{\scoregain{79.9}{\uparrow}{14.7}} & \bestscore{\scoregain{74.1}{\uparrow}{10.7}} & \bestscore{\scoregain{80.4}{\uparrow}{18.1}} & \bestscore{\scoregain{85.4}{\uparrow}{18.7}} & \bestscore{\scoregain{84.4}{\uparrow}{16.1}} & \bestscore{\scoregain{83.8}{\uparrow}{17.6}} \\
            \bottomrule
        \end{tabular*}
    \end{table*}

    \begin{table*}[t]
    \centering
    \caption{\textbf{Comparison on the five-category CoPro benchmark.}
    Precision and recall are expressed as percentages.
    Yellow highlights the best score, and green highlights the strongest baseline in each column.
    Superscripts report absolute gains in percentage points over the strongest baseline.
    Evaluation uses 5,000 validation instances, with 1,000 instances per category.}
    \label{tab:copro_category_results}
    \footnotesize
    \setlength{\tabcolsep}{2.5pt}
    \renewcommand{\arraystretch}{1.12}
    \newcommand{\coprometricwidth}{\dimexpr(\textwidth-.17\textwidth-21\tabcolsep-5\arrayrulewidth)/10\relax}
    \begin{tabular*}{\textwidth}{@{}>{\raggedright\arraybackslash}p{.17\textwidth}|
        *{4}{>{\centering\arraybackslash}p{\coprometricwidth}>{\centering\arraybackslash}p{\coprometricwidth}|}
        >{\centering\arraybackslash}p{\coprometricwidth}>{\centering\arraybackslash}p{\coprometricwidth}}
        \toprule
        \multirow{2}{*}{\textbf{Method}}
        & \multicolumn{2}{c|}{\textbf{Safe}}
        & \multicolumn{2}{c|}{\textbf{Sexual}}
        & \multicolumn{2}{c|}{\textbf{Hate}}
        & \multicolumn{2}{c|}{\textbf{Violence}}
        & \multicolumn{2}{c}{\textbf{Shocking}} \\
        \cmidrule(lr){2-3}
        \cmidrule(lr){4-5}
        \cmidrule(lr){6-7}
        \cmidrule(lr){8-9}
        \cmidrule(lr){10-11}
        & Prec (\%) & Rec (\%)
        & Prec (\%) & Rec (\%)
        & Prec (\%) & Rec (\%)
        & Prec (\%) & Rec (\%)
        & Prec (\%) & Rec (\%) \\
        \midrule
        Zero-shot
        & 48.4 & 51.2
        & 51.8 & 49.1
        & 50.9 & 55.8
        & 53.8 & 60.4
        & 56.6 & 43.9 \\
        Fixed few-shot
        & 65.6 & 68.2
        & 71.7 & 71.3
        & 65.4 & 64.7
        & 70.5 & 76.1
        & 66.2 & 59.3 \\
        Fine-tuning
        & 80.0 & 82.0
        & 84.7 & 83.2
        & 80.6 & 84.8
        & 83.1 & 85.9
        & 81.7 & 74.1 \\
        Class-RAG
        & \baselinescore{83.0} & \baselinescore{84.3}
        & 87.8 & \baselinescore{87.0}
        & \baselinescore{82.8} & \baselinescore{87.2}
        & 86.2 & \baselinescore{88.0}
        & \baselinescore{84.2} & 77.4 \\
        Llama Guard
        & 82.2 & 83.1
        & \baselinescore{88.1} & \baselinescore{87.0}
        & 81.6 & 84.3
        & \baselinescore{87.0} & 82.3
        & 82.6 & \baselinescore{84.5} \\
        \midrule
        \rowcolor{gray!12}\textbf{CaseLaw (ours)}
        & \bestscore{\scoregain{85.4}{\uparrow}{2.4}}
        & \bestscore{\scoregain{88.2}{\uparrow}{3.9}}
        & \bestscore{\scoregain{91.0}{\uparrow}{2.9}}
        & \bestscore{\scoregain{92.3}{\uparrow}{5.3}}
        & \bestscore{\scoregain{88.0}{\uparrow}{5.2}}
        & \bestscore{\scoregain{90.0}{\uparrow}{2.8}}
        & \bestscore{\scoregain{94.5}{\uparrow}{7.5}}
        & \bestscore{\scoregain{92.2}{\uparrow}{4.2}}
        & \bestscore{\scoregain{91.2}{\uparrow}{7.0}}
        & \bestscore{\scoregain{87.0}{\uparrow}{2.5}} \\
        \bottomrule
    \end{tabular*}
\end{table*}
    
\section{Experiments}

    \subsection{Experimental Setup}

    \textbf{Datasets.} We construct MFO-RedNotes (Male-Female Opposition RedNotes), a multimodal content moderation dataset collected from publicly available posts on the RedNotes platform. The dataset focuses on male-female opposition as a representative risk domain in community atmosphere governance, where harmfulness is often expressed through implicit group attribution, emotionally charged claims, gender-based antagonism, sarcastic framing, or value-laden narratives rather than explicit toxic language. MFO-RedNotes contains 50,000 public posts, each consisting of multiple content fields, including post title, main textual content, hashtags, images, videos, and a policy-dependent risk label. The risk labels are annotated by 20 expert content moderators according to the corresponding moderation policy for the male-female opposition risk domain. To model policy changes involving fine-grained exception rules, each instance is additionally annotated with an exemption indicator and an exemption rationale, both of which are used exclusively to define policy-change evaluation settings and are never provided as input to the moderation models. To support model development and evaluation, we split the dataset into 40,000 training examples, 5,000 validation examples, and 5,000 test examples. Policy revisions change the labels of 10,000 notes, or 20\% of the full benchmark, while their content and split assignments remain unchanged. The original and revised labels enable paired evaluation of moderation performance before and after a policy change. Annotation proceeds in five rounds, in each of which a randomly selected 10\% of the annotations undergo double-blind review, with reannotation required unless agreement exceeds 95\%. Appendix~\ref{app:benchmark} gives the four-level taxonomy and four expert-defined exemption rules. This dataset provides a realistic benchmark for evaluating whether moderation systems can handle multimodal, policy-sensitive, and boundary-case content under evolving governance requirements.

    CoPro~\cite{liu2024latent} is a public content safety dataset containing unsafe prompts and their corresponding safe rewritten prompts. It covers seven harmful categories. We evaluate a five-category CoPro benchmark covering Safe, Sexual, Hate, Violence, and Shocking, using a balanced validation set of 5,000 instances with 1,000 per category.

    \textbf{Baselines.} We compare CaseLaw with five moderation baselines:
    \begin{itemize}
        \item \textbf{Zero-shot.} A frozen model predicts labels from policy-informed prompts without retrieved examples.
        \item \textbf{Fixed few-shot.} Predictions use retrieved examples from a case library that stays fixed across policy changes.
        \item \textbf{Fine-tuning.} The model undergoes full-parameter fine-tuning with policy information in the training context.
        \item \textbf{Class-RAG}~\cite{chen2024class}. A generative fine-tuned classifier uses policy context and examples retrieved from an editable case library.
        \item \textbf{Llama Guard}~\cite{inan2023llama}. A general-purpose safeguard model is evaluated by mapping its safety categories to the benchmark labels.
    \end{itemize}

    \begin{figure*}[t]
        \centering
        \begin{minipage}[t]{0.325\textwidth}
            \centering
            \includegraphics[width=\textwidth]{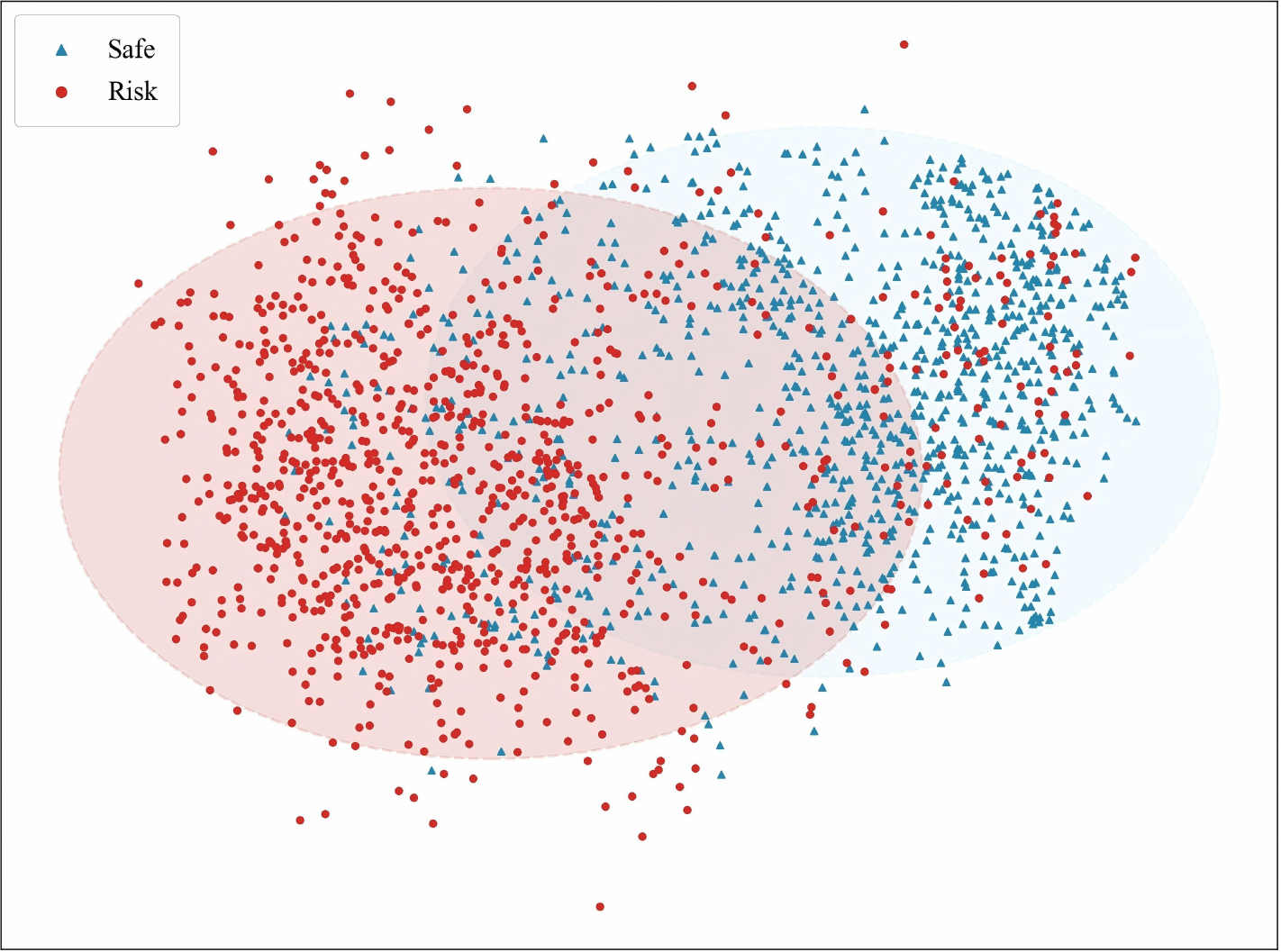}
            \par\smallskip
            \textbf{(a)}
        \end{minipage}
        \hfill
        \begin{minipage}[t]{0.325\textwidth}
            \centering
            \includegraphics[width=\textwidth]{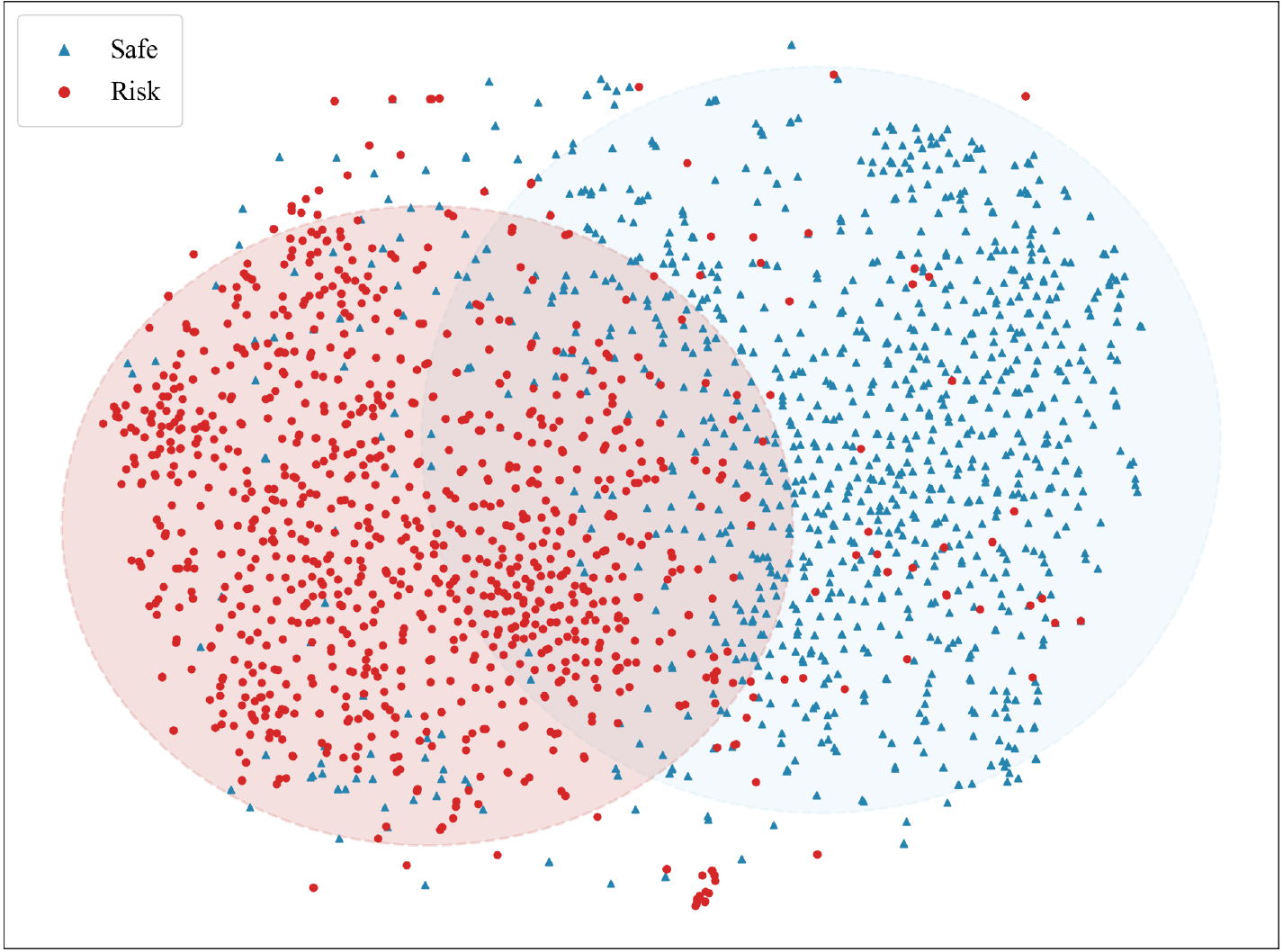}
            \par\smallskip
            \textbf{(b)}
        \end{minipage}
        \hfill
        \begin{minipage}[t]{0.325\textwidth}
            \centering
            \includegraphics[width=\textwidth]{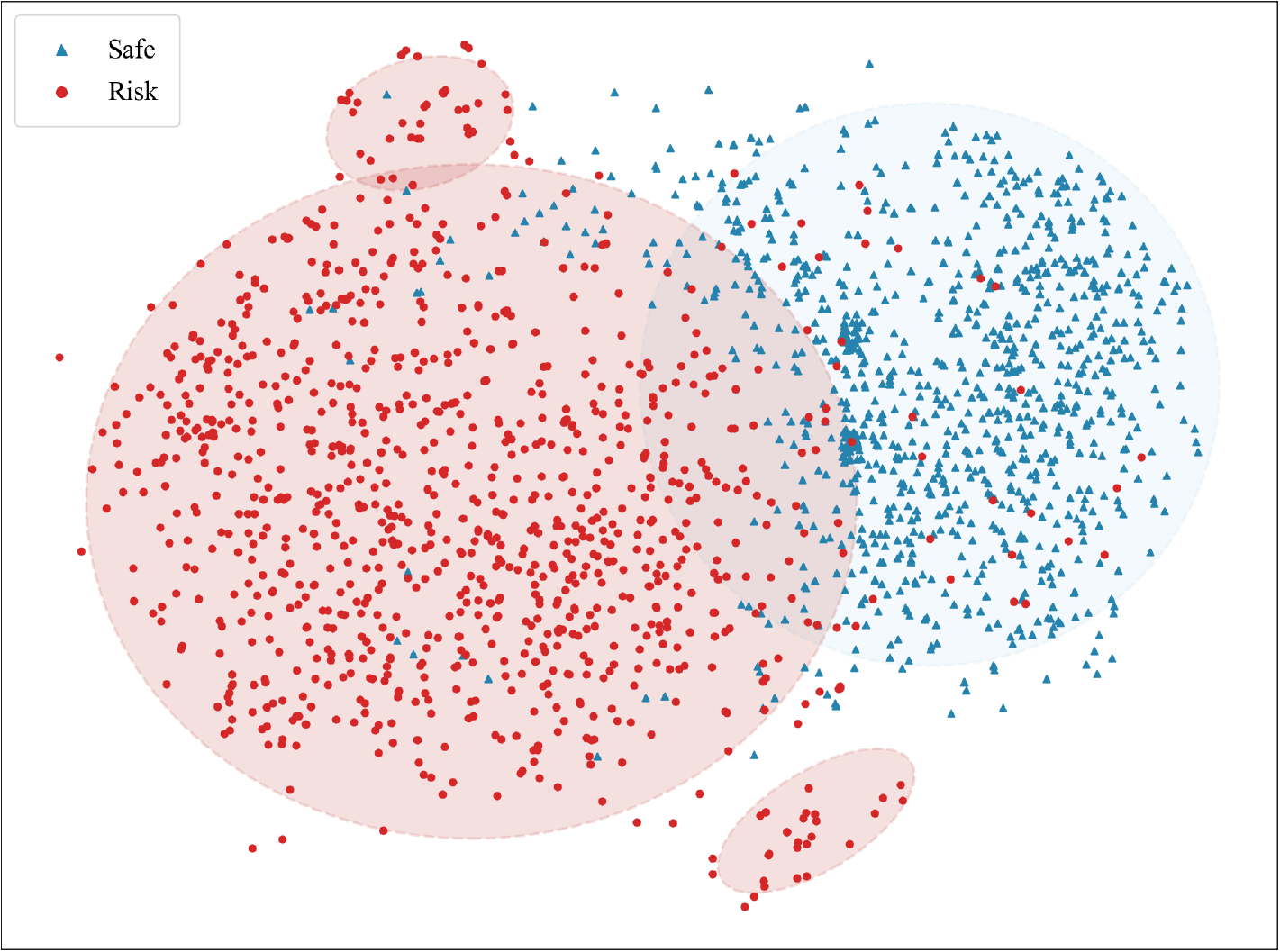}
            \par\smallskip
            \textbf{(c)}
        \end{minipage}
        \caption{Impact of structured semantic extraction on embedding distributions. We visualize 1,000 safe notes and 1,000 risk notes from MFO-RedNotes, where risk includes Mid-risk and High-risk instances. Compared with direct embedding of raw notes and summary-only representations, NoteCap produces more compact intra-class clusters and clearer inter-class separation. The concentrated risk regions further indicate that structured semantic extraction can form risk-oriented semantic anchors for more accurate downstream case retrieval.}
        \label{fig:impact-on-structured-semantic-extraction}
    \end{figure*}

    \textbf{Implementation Details.} We use Qwen3.5-35B-A3B for the Zero-shot, Fixed few-shot, and Fine-tuning language-model backbones and for the reference structured extractor. Fine-tuning uses LLaMA-Factory with full-parameter updates. Zero-shot receives policy information explicitly. CaseLaw's fixed extraction template describes the semantic schema without encoding the current policy, and precedent labels supply policy-specific supervision at inference. CaseLaw and Fixed few-shot share the same initial library. Fixed few-shot keeps its initial precedents unchanged across policy states, whereas CaseLaw updates affected case labels and metadata.

    All representation experiments use Qwen3-Embedding-8B. The NoteCap stability study uses Qwen3-Embedding-8B across all extraction variants. NoteCap training is a one-time preparation step, not a policy-update operation. CaseLaw uses the thresholded voting rule defined in Eq.~\eqref{eq:vote-count}, so it has no classification-model backbone. Llama Guard is evaluated using its original safeguard checkpoint.

    We use $k=16$ for CaseLaw in Table~\ref{tab:risk_level_results}. The experiments use a server with 32 NVIDIA H800 GPUs.

    \subsection{Performance Analysis}

    Table~\ref{tab:risk_level_results} reports class-wise precision and recall as percentages across four policy-dependent risk levels on MFO-RedNotes. Under the original policy, CaseLaw achieves the best results in all eight comparisons, reaching 85.3\%/88.3\% precision/recall on Safe and 85.3\%/90.1\% on High-risk. Its gains over the strongest baseline in each column range from 1.5 to 11.8 percentage points. These results demonstrate that CaseLaw improves both the identification of violations and the limitation of false positives.

    After policy change, CaseLaw leads all eight precision/recall comparisons, with gains of 10.7--18.7 percentage points over the strongest baseline in each column. It obtains 86.8\%/88.0\% on Safe, 79.9\%/74.1\% on Low-risk, 80.4\%/85.4\% on Mid-risk, and 84.4\%/83.8\% on High-risk. The largest margin is 18.7 percentage points in both Safe recall and Mid-risk recall, where Class-RAG reaches 69.3\% and 66.7\%. Averaged equally across the four classes, CaseLaw's precision and recall decrease from 84.6\%/83.9\% to 82.9\%/82.8\%, respectively, indicating that it retains strong overall performance under the revised policy boundaries.

    The baselines show larger declines under policy change. Fine-tuning's macro-averaged precision/recall falls from 77.2\%/76.6\% to 65.1\%/65.6\%, while Class-RAG declines from 79.5\%/78.6\% to 66.0\%/66.1\%. Fixed few-shot also degrades, with Mid-risk precision/recall falling from 64.6\%/63.4\% to 47.6\%/47.0\%. These changes highlight the difficulty of adapting to fine-grained exemptions when the surface content remains unchanged. CaseLaw’s stronger post-change performance stems from its ability to preserve stable case-level semantics while adapting the evolving mapping between these semantics and policy-dependent risk labels through case-based judging.

    Table~\ref{tab:copro_category_results} reports results across five CoPro categories. CaseLaw achieves the highest precision and recall in all ten category-level comparisons, with gains of 2.4--7.5 percentage points over the strongest baseline in each column. Its macro-averaged precision reaches 90.02\%, exceeding Class-RAG which is  84.80\% by 5.22 percentage points, while its macro-averaged recall reaches 89.94\%, surpassing Class-RAG which is 84.78\% by 5.16 percentage points. Class-RAG is the strongest baseline for both aggregate metrics. CaseLaw attains 94.5\%/92.2\% precision/recall on Violence and 91.2\%/87.0\% on Shocking, with recall gains of 4.2 and 2.5 percentage points, respectively. Its highest category-wise recall is 92.3\% on Sexual.

    \section{Ablation Studies}
    \label{sec:ablations}
    \subsection{Structured Representations}

    We study how semantic representation affects both moderation performance and its consistency across repeated runs. The MFO-RedNotes ablation uses a balanced subset of 1,000 Safe and 1,000 Unsafe notes. Direct splicing directly embeds a fixed serialization of the original note content, without generative rewriting. Summary only uses an LLM to reorganize each note into a compact semantic summary before embedding. NoteCap instead preserves risk-bearing evidence within structured semantic fields, including topic, expression type, and core claim. All representations use the same Qwen3-Embedding-8B encoder. Figure~\ref{fig:impact-on-structured-semantic-extraction} visualizes their embedding distributions with t-SNE~\cite{van2008visualizing}. Table~\ref{tab:sse_ablation} reports precision and recall averaged over five experimental runs, together with their sample variances. Appendix~\ref{app:metrics} defines the aggregation.

    \begin{table}[t]
        \centering
        \caption{\textbf{Representation ablation on MFO-RedNotes and CoPro.}
        The MFO-RedNotes subset contains 1,000 Safe and 1,000 Unsafe notes.
        Prec and Rec are means over five runs (\%); Var(P) and Var(R) are sample variances in squared percentage points (pp$^2$).}
        \label{tab:sse_ablation}
        \footnotesize
        \setlength{\tabcolsep}{2pt}
        \renewcommand{\arraystretch}{1.12}
        \resizebox{\columnwidth}{!}{%
        \begin{tabular}{@{}l|cccc|cccc@{}}
        \toprule
        \multirow{2}{*}{\textbf{Method}}
        & \multicolumn{4}{c|}{\textbf{Safe}}
        & \multicolumn{4}{c}{\textbf{Unsafe}} \\
        \cmidrule(lr){2-5}\cmidrule(lr){6-9}
        & \textbf{Prec} & \textbf{Rec} & \textbf{Var(P)} & \textbf{Var(R)}
        & \textbf{Prec} & \textbf{Rec} & \textbf{Var(P)} & \textbf{Var(R)} \\
        \midrule
        \multicolumn{9}{@{}l}{\textit{MFO-RedNotes}} \\
        Direct splicing
        & 66.0 & 68.8 & 0.00 & 0.00 & 67.4 & 64.6 & 0.00 & 0.00 \\
        Summary only
        & 73.0 & 74.5 & 0.09 & 0.09 & 74.0 & 72.5 & 0.09 & 0.09 \\
        \textbf{NoteCap}
        & \textbf{85.6} & \textbf{88.0} & 0.00 & 0.00 & \textbf{87.7} & \textbf{85.2} & 0.00 & 0.00 \\
        \midrule
        \multicolumn{9}{@{}l}{\textit{CoPro}} \\
        Direct splicing
        & 61.0 & 66.9 & 0.00 & 0.00 & 71.3 & 65.8 & 0.00 & 0.00 \\
        Summary only
        & 68.2 & 73.3 & 0.11 & 0.12 & 77.3 & 72.6 & 0.07 & 0.12 \\
        \textbf{NoteCap}
        & \textbf{85.4} & \textbf{88.4} & 0.00 & 0.00 & \textbf{90.5} & \textbf{87.9} & 0.00 & 0.00 \\
        \bottomrule
        \end{tabular}%
        }
    \end{table}

    \begin{table*}[t]
    \centering
    \caption{\textbf{Five-run extraction stability and moderation performance on the MFO-RedNotes validation split.}
    Evaluation uses the original policy and voting depth $k=16$.
    Unsafe combines Low-risk, Mid-risk, and High-risk.
    Top5-diff is the percentage of notes whose top-five retrieved case sets change across five runs.
    Precision and recall are five-run means (\%); their sample variances are measured in squared percentage points (pp$^2$).}
    \label{tab:notecap-stability}
    \footnotesize
    \setlength{\tabcolsep}{2.5pt}
    \renewcommand{\arraystretch}{1.15}
    \newcommand{\stabilitymetricwidth}{\dimexpr(\textwidth-.19\textwidth-20\tabcolsep-2\arrayrulewidth)/10\relax}
    \begin{tabular*}{\textwidth}{@{}>{\raggedright\arraybackslash}p{.19\textwidth}|
        *{5}{>{\centering\arraybackslash}p{\stabilitymetricwidth}}|
        *{5}{>{\centering\arraybackslash}p{\stabilitymetricwidth}}@{}}
    \toprule
    \multirow{2}{*}{\textbf{Extractor}}
    & \multicolumn{5}{c|}{\textbf{Safe}}
    & \multicolumn{5}{c}{\textbf{Unsafe}} \\
    \cmidrule(lr){2-6}\cmidrule(lr){7-11}
    & Top5-diff & Prec & Rec & Var(P) & Var(R)
    & Top5-diff & Prec & Rec & Var(P) & Var(R) \\
    \midrule
    Gemini-3.1-Pro & 16.8 & 77.5 & 84.4 & 0.20 & 0.22 & 12.0 & 94.6 & 91.8 & 0.02 & 0.05 \\
    GPT-6.1-sol & 11.6 & 80.0 & 86.2 & 0.12 & 0.13 & 7.6 & 95.3 & 92.8 & 0.01 & 0.03 \\
    Qwen-3.8-Max & 22.6 & 75.5 & 82.9 & 0.31 & 0.53 & 17.5 & 94.1 & 91.0 & 0.05 & 0.08 \\
    \midrule
    \rowcolor{gray!12}\textbf{NoteCap} & \textbf{0.0} & \textbf{85.3} & \textbf{89.0} & \textbf{0.00} & \textbf{0.00} & \textbf{0.0} & \textbf{96.3} & \textbf{94.9} & \textbf{0.00} & \textbf{0.00} \\
    \bottomrule
    \end{tabular*}
    \end{table*}
    
    Figure~\ref{fig:impact-on-structured-semantic-extraction} shows substantial overlap among raw-note representations, improved semantic grouping after summarization, and more compact risk-oriented regions with NoteCap. Table~\ref{tab:sse_ablation} distinguishes representation quality from run-to-run stability. Direct splicing yields the lowest precision and recall on both datasets despite producing stable results, whereas the Summary method improves both metrics but exhibits fluctuations attributable to model instability. NoteCap further improves precision and recall for both the Safe and Unsafe classes, with gains over Summary only of 12.6--13.7 percentage points on MFO-RedNotes and 13.2--17.2 percentage points on CoPro. Across five runs, NoteCap reports precision and recall variances of 0.00~pp$^2$ for both classes on both datasets. These results support NoteCap's ability to concentrate moderation-relevant semantics while maintaining stable aggregate performance under the evaluated configurations.

    \subsection{Voting Depth}
\begin{figure}[ht]
    \centering
    \includegraphics[width=\columnwidth]{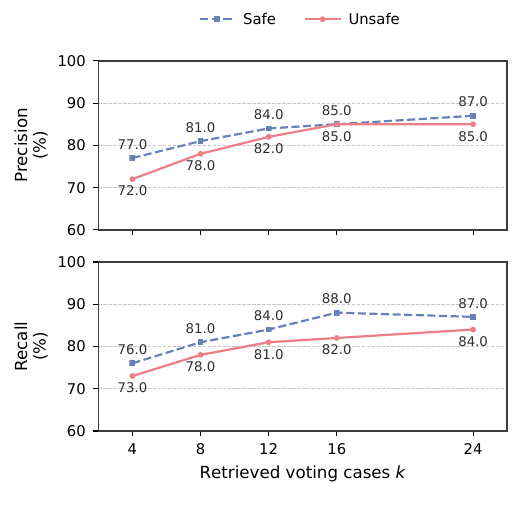}
    \caption{Voting-depth ablation under the original-policy setting of Table~\ref{tab:risk_level_results}, with all other experimental conditions fixed. Precision (top) and recall (bottom) are expressed as percentages. Blue dashed squares denote Safe and pink solid circles denote Unsafe.}
    \label{fig:topk-scaling}
    \end{figure}

    To isolate the effect of voting depth, we follow the original-policy setting of Table~\ref{tab:risk_level_results}, holding the evaluation split, extractor, encoder, case library, case-specific thresholds, and voting rule fixed while varying $k\in\{4,8,12,16,24\}$. Figure~\ref{fig:topk-scaling} summarizes performance as Safe and Unsafe, with Low-risk, Mid-risk, and High-risk grouped into Unsafe for reporting. From $k=4$ to $k=16$, Safe precision/recall increases from 77\%/76\% to 85\%/88\%, and Unsafe precision/recall increases from 72\%/73\% to 85\%/82\%. At $k=24$, these values become 87\%/87\% and 85\%/84\%, respectively. Changes beyond $k=16$ remain within two percentage points, indicating diminishing marginal gains.

    \subsection{NoteCap Stability}
    We assess extraction stability through precedent retrieval and downstream moderation on the 5,000-note MFO-RedNotes validation split under the original policy. Gemini-3.1-Pro, GPT-6.1-sol, Qwen-3.8-Max, and NoteCap each generate five structured representations per note with a fixed extraction prompt. All extractors share the Qwen3-Embedding-8B encoder, frozen case library, and thresholded voting configuration with $k=16$. Unsafe combines Low-risk, Mid-risk, and High-risk. Top5-diff measures the percentage of notes whose top-five retrieved case sets change across runs, ignoring order. Table~\ref{tab:notecap-stability} reports this rate alongside five-run precision/recall means and sample variances.

    NoteCap achieves 85.3\%/89.0\% precision/recall on Safe and 96.3\%/94.9\% on Unsafe. Compared with GPT-6.1-sol, the strongest baseline across the reported metrics, its precision/recall gains are 5.3/2.8 percentage points on Safe and 1.0/2.1 percentage points on Unsafe. NoteCap also reports 0.0\% Top5-diff for both classes, compared with 11.6\%/7.6\% for GPT-6.1-sol, and precision/recall variances of 0.00~pp$^2$. These results demonstrate improved moderation quality alongside stable top-five retrieval and aggregate performance across the five runs.

    \section{Future Analysis}
    \label{sec:future-analysis}
    This section presents four key findings from the practical application of CaseLaw, which help guide case library design.

    \textbf{Larger case libraries improve moderation performance.} Quality-controlled expansion broadens coverage of rare risk patterns and fine-grained policy boundaries, thereby enhancing the stability and robustness of the judge. These benefits stem from broader coverage of distinct moderation-relevant semantics. 

    \textbf{Increasing $k$ improves robustness, but the gains follow a scaling law.} Figure~\ref{fig:topk-scaling} reveals a saturating empirical scaling law for voting depth over the evaluated range: increasing $k$ initially improves precision and recall, with diminishing gains beyond $k=16$. More retrieved precedents provide additional evidence, but their marginal benefit diminishes once the voting neighborhood contains sufficient relevant cases.

    \textbf{Case counts across risk labels should remain balanced within the library.} Otherwise, frequently represented labels can dominate local votes because of storage frequency rather than semantic relevance. Balanced admission and stratified sampling can address this bias.

    \textbf{Cases in the library should remain semantically distinct, avoiding excessive similarity among cases.} Case admission should favor distinct evidence patterns and decision boundaries. Here semantic independence means limited redundancy. Many nearly identical cases can amplify correlated annotation errors and overwhelm a vote without adding meaningful coverage.

    \section{Conclusion} 
        In this paper, we presented CaseLaw, a policy-decoupled framework for training-free content moderation. By separating stable semantic representation from dynamic policy interpretation, CaseLaw enables moderation systems to adapt to evolving policies through updates to an external semantic case library rather than through rule rewriting, model retraining, or redeployment. Its NoteCap extractor preserves risk-bearing evidence while organizing semantic attributes; retrieved precedent labels then determine the outcome through thresholded voting. We further introduced MFO-RedNotes, a multimodal benchmark for the male-female opposition risk domain, and evaluated CaseLaw under its original and revised policies. These experiments demonstrate that CaseLaw yields consistent gains over zero-shot, fixed few-shot, and fine-tuning baselines, while ablation studies validate the effectiveness of structured semantic extraction in forming clearer and more policy-aligned semantic anchors. These findings indicate that case-library-driven policy adaptation offers a practical, scalable, and transparent direction for content moderation under rapidly evolving governance requirements in operationally realistic settings.

\section{Limitations}
CaseLaw externalizes policy knowledge rather than removing policy dependence: performance remains sensitive to the accuracy, coverage, balance, and freshness of the case library. Comparisons with general safeguard models are sensitive to the mapping between their native safety categories and the benchmark's policy-dependent labels.

\newpage
\bibliography{aaai2027}
\clearpage
\appendix
\setcounter{secnumdepth}{1}
\section{Structured Semantic Refinement Reduces Moderation Uncertainty}

    This section provides an information-theoretic perspective on CaseLaw, showing that structured semantic refinement can reduce moderation uncertainty by conditioning decisions on policy-relevant precedent clusters beyond coarse classifier regions. We consider a binary specialization of the policy-dependent label space $\mathcal{Y}_t$, where $x$ denotes the input content and $y^{(t)}\in\{0,1\}$ denotes the moderation label. In the male--female opposition domain, $y^{(t)}=1$ means violation and $y^{(t)}=0$ means pass. A conventional classifier learns a posterior $q_\Theta(y^{(t)}\mid x)$ through cross-entropy training. Let $R$ denote a coarse semantic region induced by the classifier representation. In contrast, our case-based framework first transforms content into the structured semantic representation $a_x=\mathcal{S}_{\theta}(x)$, and then retrieves a fine-grained semantic precedent cluster $\mathcal{N}$ induced by $\mathcal{N}_k(x)$.
    
    Intuitively, $R$ is what the classifier can distinguish, while $\mathcal{N}$ is what structured semantic reasoning can further refine.
    
    \subsection{Entropy Reduction by Semantic Refinement}
    
    \begin{proposition}[Entropy Reduction by Semantic Refinement]
    If the structured semantic cluster $\mathcal{N}$ provides additional information about the moderation label $y^{(t)}$ beyond the classifier's coarse region $R$, then conditioning on $\mathcal{N}$ reduces moderation uncertainty:
    \begin{equation}
    H(y^{(t)}\mid R,\mathcal{N})<H(y^{(t)}\mid R).
    \end{equation}
    More generally,
    \begin{equation}
    H(y^{(t)}\mid R)-H(y^{(t)}\mid R,\mathcal{N})=I(y^{(t)};\mathcal{N}\mid R)\geq 0.
    \end{equation}
    The inequality is strict when
    \begin{equation}
    I(y^{(t)};\mathcal{N}\mid R)>0.
    \end{equation}
    \end{proposition}
    
    \begin{proof}
    \renewcommand{\qedsymbol}{}
    Fix a coarse classifier region $R=r$. Suppose structured semantic reasoning refines this region into $m$ fine-grained clusters:
    \begin{equation}
    R=r=\mathcal{N}_1\cup \mathcal{N}_2\cup\cdots\cup \mathcal{N}_m.
    \end{equation}
    Let
    \begin{equation}
    p_j=P(y^{(t)}=1\mid R=r,\mathcal{N}=j)
    \end{equation}
    be the violation rate of the $j$-th semantic cluster, and let
    \begin{equation}
    \lambda_j=P(\mathcal{N}=j\mid R=r)
    \end{equation}
    be its mixture weight inside the coarse region. Then
    \begin{equation}
    \sum_{j=1}^m \lambda_j=1.
    \end{equation}
    By the law of total probability, the violation rate of the coarse classifier region is
    \begin{equation}
    q_r=P(y^{(t)}=1\mid R=r)=\sum_{j=1}^{m}\lambda_jp_j.
    \end{equation}
    The binary entropy function is
    \begin{equation}
    h(p)=-p\log p-(1-p)\log(1-p).
    \end{equation}
    Since $h(p)$ is concave, Jensen's inequality gives
    \begin{equation}
    h\left(\sum_{j=1}^{m}\lambda_jp_j\right)
    \geq
    \sum_{j=1}^{m}\lambda_jh(p_j).
    \end{equation}
    Therefore,
    \begin{equation}
    h(q_r)
    \geq
    \sum_{j=1}^{m}\lambda_jh(p_j).
    \end{equation}
    Equivalently,
    \begin{equation}
    H(y^{(t)}\mid R=r)
    \geq
    H(y^{(t)}\mid R=r,\mathcal{N}).
    \end{equation}
    Averaging over all coarse regions $R$, we obtain
    \begin{equation}
    H(y^{(t)}\mid R)-H(y^{(t)}\mid R,\mathcal{N})
    =
    I(y^{(t)};\mathcal{N}\mid R)
    \geq 0.
    \end{equation}
    Thus, structured semantic refinement cannot increase uncertainty. It strictly reduces uncertainty whenever the semantic cluster $\mathcal{N}$ contains policy-relevant information not captured by the classifier's coarse region $R$.
    \end{proof}
    
\begin{table*}[t]
    \centering
    \small
    \setlength{\tabcolsep}{6pt}
    \renewcommand{\arraystretch}{1.18}
    \begin{tabular}{@{}>{\raggedright\arraybackslash}m{2.4cm}m{14.2cm}@{}}
    \toprule
    \textbf{Risk Level} &
    \textbf{Typical Manifestation} \\
    \midrule

    \textbf{Safe}
    &
    No obvious risk of gender-based antagonism. The content is typically neutral, factual, or rights-oriented, and respects diverse viewpoints without inflammatory framing.
    \\

    \textbf{Low Risk}
    &
    The content may involve sensitive gender-related topics but remains restrained in tone. It usually presents controversial opinions in a moderate manner without explicit hostility, group attribution, or provocative escalation.
    \\

    \textbf{Medium Risk}
    &
    The content shows a clear antagonistic tendency but does not reach an extreme or explicit violation. It often uses oppositional framing, implicit group attribution, emotionally charged claims, or partially matches policy-defined risky patterns.
    \\

    \textbf{High Risk}
    &
    The content exhibits strong antagonism or directly violates the moderation policy. Typical manifestations include original-sin rhetoric, extreme demands, explicit abuse, hate speech, deliberate provocation, or intensified gender-based opposition.
    \\

    \bottomrule
    \end{tabular}
    \caption{Risk-level taxonomy for the male-female opposition moderation domain. Each level reflects increasing severity of gender-based antagonism and is defined according to expert-designed content-safety policies used for MFO-RedNotes.}
    \label{tab:risk_classification}

\end{table*}

    \subsection{Disagreement View}
    
    The same result can be expressed using pairwise label disagreement. For a binary cluster with violation rate $p$, define
    \begin{equation}
    D(p)=2p(1-p).
    \end{equation}
    This is the probability that two randomly sampled cases from the same region have different labels.
    
    For the coarse region $R=r$, the disagreement is
    \begin{equation}
    D(q_r)=2q_r(1-q_r).
    \end{equation}
    For the refined semantic clusters, the average disagreement is
    \begin{equation}
    \sum_{j=1}^{m}\lambda_jD(p_j).
    \end{equation}
    Then
    \begin{equation}
    D(q_r)-\sum_{j=1}^{m}\lambda_jD(p_j)
    =
    2\operatorname{Var}_{\lambda}(p_j)
    \geq 0.
    \end{equation}
    Indeed,
    \begin{align}
    D(q_r)-\sum_{j=1}^{m}\lambda_jD(p_j)
    &=\left(2q_r-2q_r^2\right)
    -
    \sum_{j=1}^{m}\lambda_j\left(2p_j-2p_j^2\right) \\
    &=2\sum_{j=1}^{m}\lambda_jp_j^2-2q_r^2 \\
    &=2\operatorname{Var}_{\lambda}(p_j).
    \end{align}
    Therefore, when a classifier collapses semantic modes with different violation rates into one coarse region, it necessarily increases local label disagreement. Structured semantic case reasoning reduces this disagreement by separating those modes into policy-aligned precedent clusters.
    
    \subsection{Interpretation}
    
    The classifier observes a coarse mixture region $R$, so the cross-entropy optimal prediction is the mixture posterior
    \begin{equation}
    P(y^{(t)}=1\mid R).
    \end{equation}
    Structured case reasoning observes a finer semantic cluster $\mathcal{N}$, so it estimates
    \begin{equation}
    P(y^{(t)}=1\mid R,\mathcal{N}).
    \end{equation}
    The benefit depends on whether retrieved clusters capture additional policy-relevant information. Modern foundation models can construct the structured semantic representation $a_x=\mathcal{S}_{\theta}(x)$ that is aligned with moderation policy factors, such as target group, generalization, essentialist attribution, derogation, incitement, quotation, and context.
    
    Thus, the key difference is:
    \begin{align}
    \text{classification head:} \quad & P(y^{(t)}\mid R), \\
    \text{structured case reasoning:} \quad & P(y^{(t)}\mid R,\mathcal{N}).
    \end{align}
    When
    \begin{equation}
    I(y^{(t)};\mathcal{N}\mid R)>0,
    \end{equation}
    the structured semantic cluster provides additional moderation information, leading to lower entropy and lower label disagreement.

    \begin{figure*}[t]
        \centering
        \begin{minipage}[t]{0.325\textwidth}
            \centering
            \includegraphics[width=\textwidth]{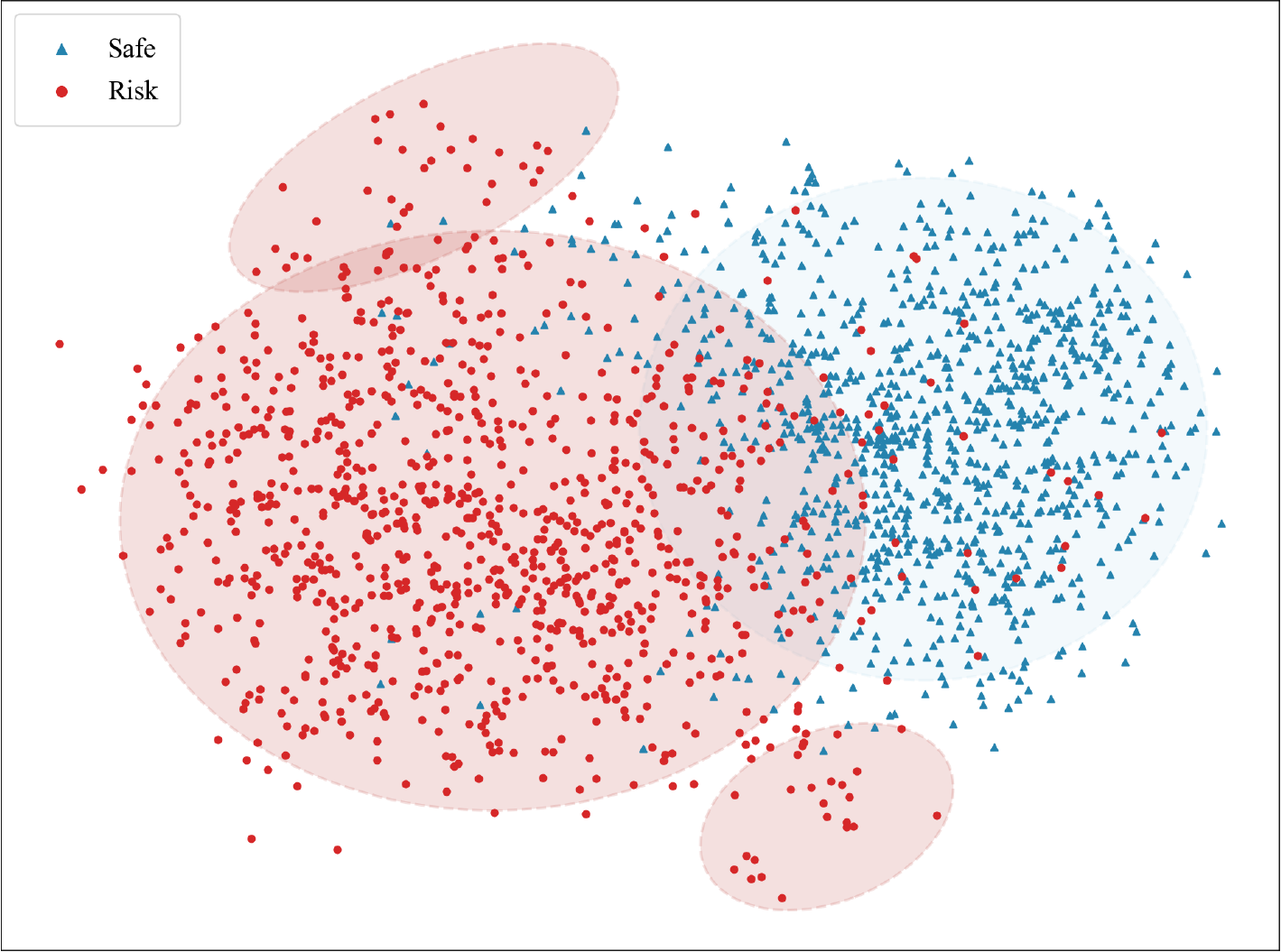}
            \par\smallskip
            \textbf{(a)}
        \end{minipage}
        \hfill
        \begin{minipage}[t]{0.325\textwidth}
            \centering
            \includegraphics[width=\textwidth]{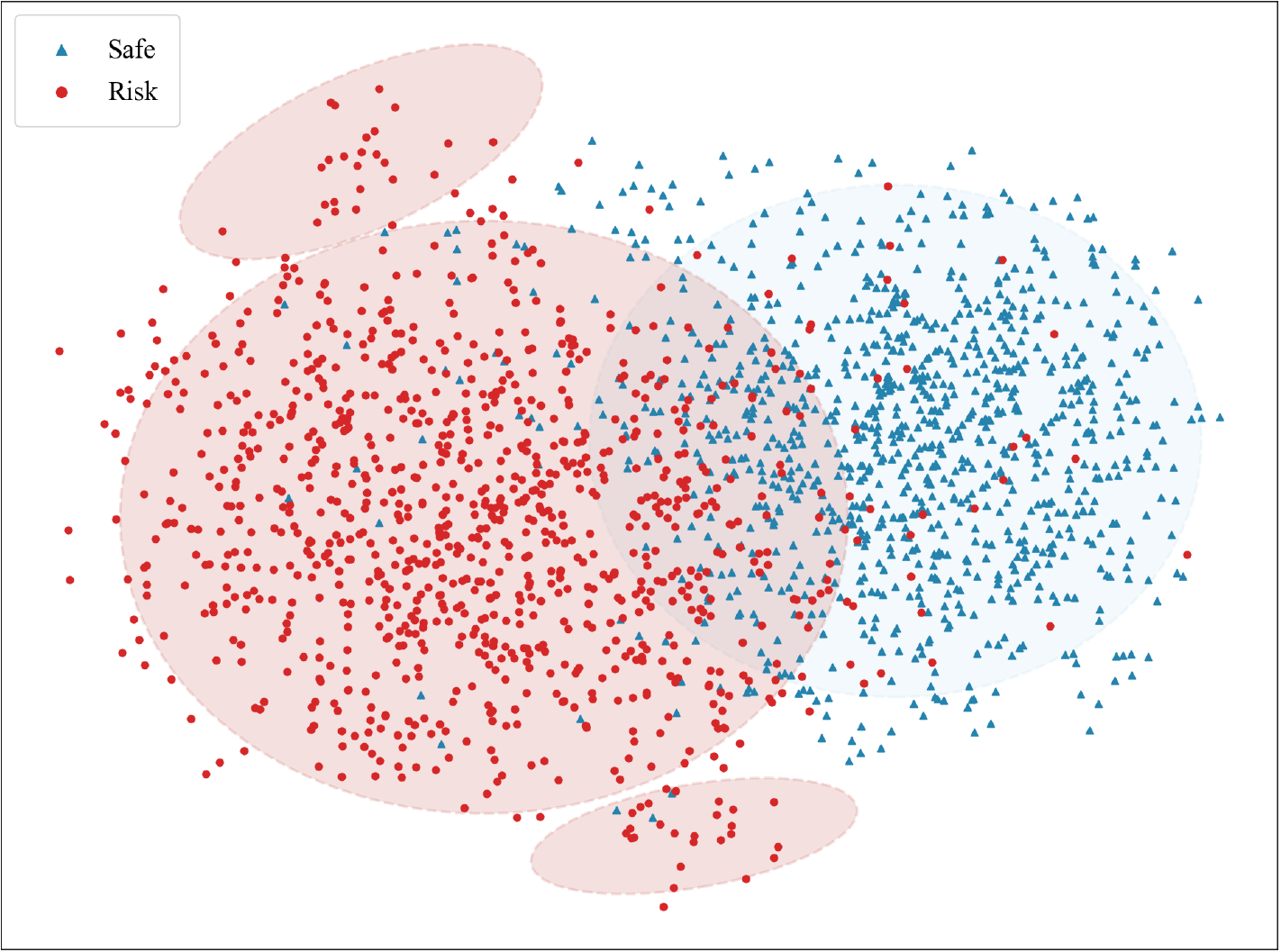}
            \par\smallskip
            \textbf{(b)}
        \end{minipage}
        \hfill
        \begin{minipage}[t]{0.325\textwidth}
            \centering
            \includegraphics[width=\textwidth]{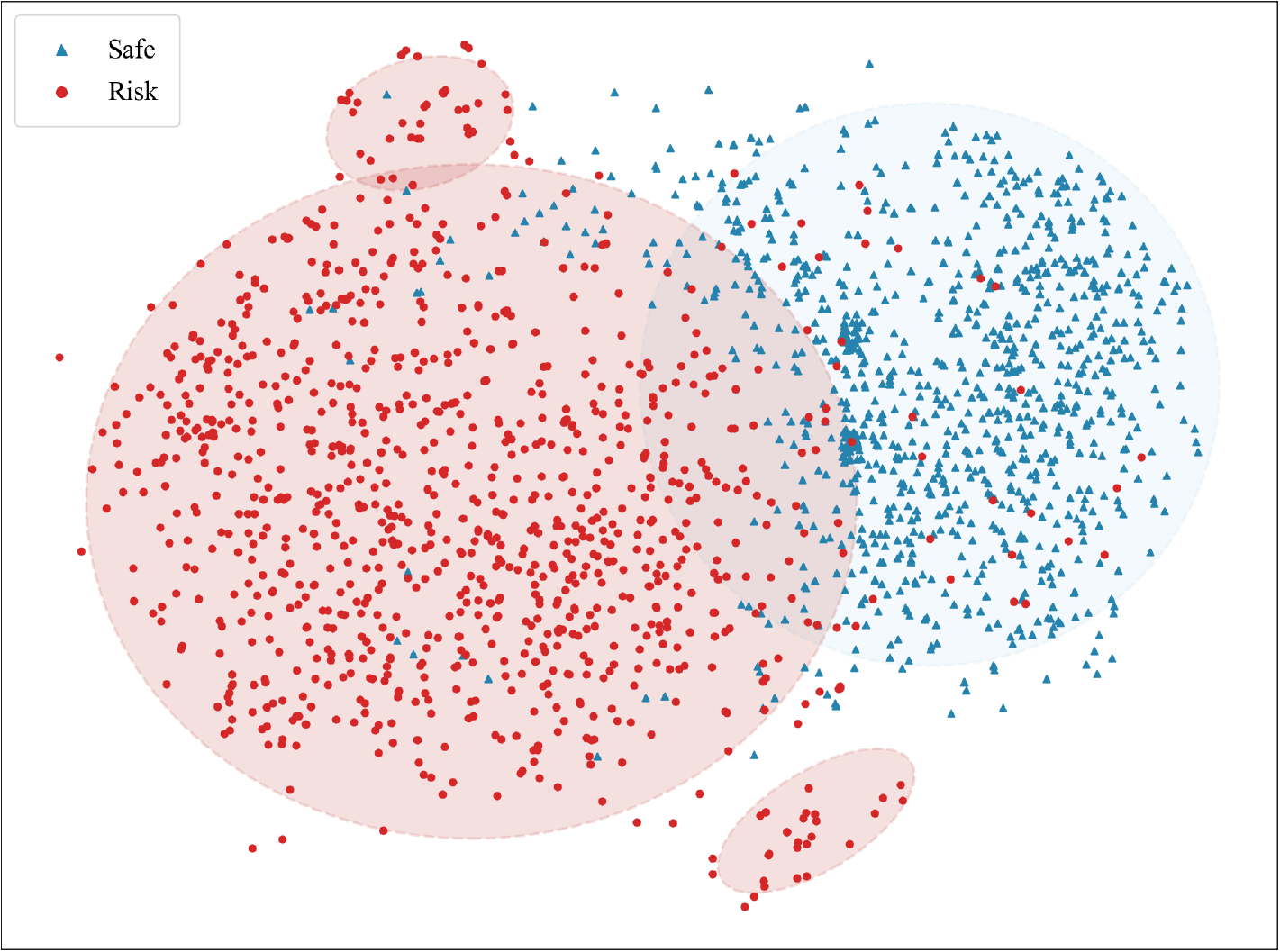}
            \par\smallskip
            \textbf{(c)}
        \end{minipage}
        \caption{Impact of different structured semantic extraction backbones on embedding distributions. We visualize the representations produced by (a) InternVL3-14B-Instruct, (b) Llama-4-Scout-17B, and (c) NoteCap on MFO-RedNotes. NoteCap produces more compact intra-class clusters and clearer safe--risk separation, indicating better formation of risk-oriented semantic anchors for downstream retrieval.}
        \label{fig:impact-on-model}
    \end{figure*}

\section{Details of MFO-RedNotes benchmark}
\label{app:benchmark}
MFO-RedNotes contains 50,000 multimodal notes partitioned into 40,000 training, 5,000 validation, and 5,000 test instances.

\subsection{Exemption Details}

    To evaluate policy adaptation under fine-grained boundary changes, we introduce exemption rules to simulate a realistic policy-change scenario in MFO-RedNotes.  Prior studies on online sexism and gender-related language show that such expressions can range from explicit sexist hate speech to subtle ambivalent sexism, and often require fine-grained taxonomies for reliable interpretation~\cite{talat2016hateful,jha2017does,lei2024systematic}. Before exemption rules are applied, some cases containing sensitive gender-related expressions may be assigned higher risk levels due to their apparent association with male-female opposition. After the introduction of exemption rules, cases satisfying expert-defined exemption conditions are downgraded to lower risk levels, reflecting a revised policy boundary rather than a change in surface content. These exemption rules are formulated by industry content-safety experts and are used only to construct the policy-change evaluation setting; the exemption indicator and exemption rationale are never provided as input to the evaluated moderation methods.

    The exemption rules cover the following four types:

    \begin{itemize}
        \item \textbf{Non-negative conflict or emotion.} Content can be exempted when it expresses non-antagonistic emotional experiences, such as pain caused by missing someone, regret over a missed relationship, or frustration about not finding a partner.

        \item \textbf{Positive viewpoints or constructive cases.} Content can be exempted when it promotes constructive gender-related perspectives, such as encouraging friendly problem-solving between genders or sharing successful cases where both sides reach mutual understanding.

        \item \textbf{Educational legal or risk-prevention content.} Content can be exempted when it only provides legal knowledge or risk-prevention information, as long as it does not associate the discussion with real-world social events, social phenomena, or public controversies.

        \item \textbf{Individual-level factual discussion.} Content can be exempted when it discusses only a specific individual, does not generalize the description to a broader gender group, and the individual-level description is factual.
    \end{itemize}

    These rules create a challenging policy-change setting because the surface expressions of many cases remain unchanged, while their effective risk labels may shift after the introduction of exemption conditions. Therefore, this setting allows us to examine whether moderation systems can capture subtle policy-boundary revisions rather than merely relying on lexical risk cues.

\subsection{Risk Level Definitions}

    Table~\ref{tab:risk_classification} summarizes the risk-level taxonomy used in MFO-RedNotes. The annotation policy is formulated by industry content-safety experts for the male-female opposition risk domain, where the key challenge lies in distinguishing neutral discussion, controversial but restrained expression, implicit antagonism, and explicit policy-violating hostility. Following this policy, each post is assigned one of four risk levels: Safe, Low Risk, Medium Risk, or High Risk, reflecting increasing degrees of antagonistic framing, group opposition, rhetorical escalation, and potential harm.

    The risk labels are annotated by 20 senior content moderators according to the corresponding domain-specific moderation policy. The annotation process is conducted in five submission rounds; for each round, 10\% of the submitted annotations are randomly sampled for double-blind quality inspection. A batch is accepted only when the annotation agreement exceeds 95\%; otherwise, the corresponding instances are re-annotated until the agreement requirement is satisfied. This procedure ensures that the final labels are not only policy-dependent but also reliable for evaluating moderation systems under fine-grained and boundary-sensitive risk definitions.

\section{Experiments on More Datasets}
\label{app:more-datasets}

To further examine the generality of CaseLaw beyond multimodal note moderation, we evaluate it on textual comments from The Big Ban Theory (TBBT)~\cite{cerulli2026big}, a dataset of user activity before and after moderation interventions. We use the dataset's existing labels to evaluate three subreddit subsets, \textit{braincels}, \textit{theredpill}, and \textit{chapotraphouse}, across distinct comment-moderation settings.

We apply CaseLaw's structured semantic extraction, precedent retrieval, and thresholded voting pipeline, comparing it with Few-shot, Class-RAG, and Llama Guard. Table~\ref{tab:tbbt_results} reports precision and recall for each subset.

\begin{table}[t]
    \centering
    \caption{\textbf{Illustrative moderation performance on The Big Ban Theory benchmark.}
    Precision and recall are expressed as percentages for three subreddit subsets.
    Superscripts indicate gains in percentage points over the strongest baseline in each column.}
    \label{tab:tbbt_results}
    \footnotesize
    \setlength{\tabcolsep}{3pt}
    \renewcommand{\arraystretch}{1.12}
    \resizebox{\columnwidth}{!}{%
    \begin{tabular}{@{}l|cc|cc|cc@{}}
        \toprule
        \multirow{2}{*}{\textbf{Method}}
        & \multicolumn{2}{c|}{\textbf{braincels}}
        & \multicolumn{2}{c|}{\textbf{theredpill}}
        & \multicolumn{2}{c}{\textbf{chapotraphouse}} \\
        \cmidrule(lr){2-3}
        \cmidrule(lr){4-5}
        \cmidrule(lr){6-7}
        & Prec (\%) & Rec (\%)
        & Prec (\%) & Rec (\%)
        & Prec (\%) & Rec (\%) \\
        \midrule
        Few-shot
        & 68.3 & 69.8
        & 64.9 & 69.2
        & 70.6 & 67.1 \\
        Class-RAG
        & 83.7 & 84.6
        & 81.9 & 83.2
        & 78.2 & 79.3 \\
        Llama Guard
        & 83.6 & 85.8
        & 80.6 & 87.4
        & 85.5 & 80.3 \\
        \midrule
        \textbf{CaseLaw (ours)}
        & \scoregain{87.9}{\uparrow}{4.2}
        & \scoregain{90.3}{\uparrow}{4.5}
        & \scoregain{86.7}{\uparrow}{4.8}
        & \scoregain{92.1}{\uparrow}{4.7}
        & \scoregain{89.8}{\uparrow}{4.3}
        & \scoregain{90.5}{\uparrow}{10.2} \\
        \bottomrule
    \end{tabular}%
    }
\end{table}

CaseLaw achieves the highest precision and recall in all six comparisons, reaching 87.9\%/90.3\% on \textit{braincels}, 86.7\%/92.1\% on \textit{theredpill}, and 89.8\%/90.5\% on \textit{chapotraphouse}. The gains over the strongest baselines range from 4.2 to 10.2 percentage points, with the largest improvement in \textit{chapotraphouse} recall over Llama Guard. These results support the applicability of case-based moderation beyond multimodal notes to textual comments across different subreddit communities.

\section{Comparison of NoteCap with Open-Source Models}

NoteCap serves as CaseLaw's structured semantic extractor. The extracted representations support precedent retrieval, while the final risk label is determined by the thresholded voting rule in the main method. When comparing extraction backbones, the embedding model~\cite{qwen3embedding}, retrieval procedure, case library, case-specific thresholds, and voting rule remain fixed, isolating the representation supplied to the voting pipeline.

Figure~\ref{fig:impact-on-model} compares the embedding distributions produced by InternVL3-14B-Instruct~\cite{opengvlab2025internvl3_14b_instruct}, Llama-4-Scout-17B~\cite{meta2025llama4scout}, and NoteCap in panels (a)--(c), respectively. Relative to the two general-purpose open-source extractors, NoteCap exhibits tighter intra-class clustering and clearer separation between safe and risk content. These projected distributions provide qualitative evidence of NoteCap's stronger semantic discriminability, indicating that its structured representations concentrate risk-relevant cues while reducing cross-class overlap. This organization yields more discriminative semantic anchors for precedent retrieval and subsequent thresholded voting.

\section{Details of Further Analysis}
\label{app:further-analysis}
\label{app:library-scale}
\label{app:balance}
\label{app:independence}

\textbf{Case-Library Scale.} We vary the number of training-split cases from 1,000 to 40,000 while holding label proportions, evaluation queries, and retrieval settings fixed. Table~\ref{tab:library-scale} shows consistent improvements in all three metrics: macro-precision increases from 70.3\% to 84.3\%, macro-recall from 60.7\% to 83.6\%, and macro-F1 from 64.0\% to 83.8\%. The 19.8-percentage-point gain in macro-F1 indicates that broader precedent coverage improves moderation performance.
\begin{table}[t]
\centering\small
\resizebox{\columnwidth}{!}{%
\begin{tabular}{rccc}
\toprule
Cases & Macro-P & Macro-R & Macro-F1 \\
\midrule
1,000 & 70.3 & 60.7 & 64.0 \\
5,000 & 75.4 & 70.0 & 72.3 \\
20,000 & 79.9 & 77.3 & 78.4 \\
40,000 & 84.3 & 83.6 & 83.8 \\
\bottomrule
\end{tabular}%
}
\caption{\textbf{Effect of case-library size on moderation performance.}
Label proportions, evaluation queries, and retrieval settings are fixed.
All metrics are percentages. Macro-F1 averages class-wise F1 scores.}
\label{tab:library-scale}
\end{table}

\textbf{Label Balance in the Case Library.} Based on the 40,000-case reference configuration, we vary the ratio between the largest and smallest label counts while keeping library size, evaluation queries, and retrieval settings fixed. As the ratio increases from 1:1 to 8:1, macro-F1 decreases from 83.8\% to 69.3\%, and worst-label recall falls from 76.6\% to 51.4\% (Table~\ref{tab:label-balance}), corresponding to declines of 14.5 and 25.2 percentage points. The sharper reduction in worst-label recall highlights the importance of balanced precedent support for maintaining performance across labels.
\begin{table}[t]
\centering\small
\resizebox{\columnwidth}{!}{%
\begin{tabular}{lcc}
\toprule
Largest:smallest & Macro-F1 & Worst-label recall \\
\midrule
1:1 & 83.8 & 76.6 \\
2:1 & 81.8 & 71.9 \\
4:1 & 76.5 & 63.7 \\
8:1 & 69.3 & 51.4 \\
\bottomrule
\end{tabular}%
}
\caption{\textbf{Effect of label imbalance in the case library.}
Library size, evaluation queries, and retrieval settings are fixed.
The balanced condition bases on the 40,000-case reference configuration in Table~\ref{tab:library-scale}.
All metrics are percentages.}
\label{tab:label-balance}
\end{table}

\textbf{Semantic Independence and Case Quality.} Based on the same 40,000-case reference configuration, we vary the fraction of cases belonging to near-duplicate semantic clusters while holding label proportions, evaluation queries, and retrieval settings fixed. As the near-duplicate share increases from 0\% to 75\%, the number of semantic clusters decreases from 40,000 to 17,240, while macro-F1 falls from 83.8\% to 69.5\%, a decline of 14.3 percentage points (Table~\ref{tab:semantic-independence}). These results associate reduced semantic diversity with lower moderation performance and support admitting cases that contribute distinct evidence patterns.
\begin{table}[t]
\centering\small
\resizebox{\columnwidth}{!}{%
\begin{tabular}{lcc}
\toprule
Near-duplicate share & Clusters & Macro-F1 \\
\midrule
0\% & 40,000 & 83.8 \\
25\% & 33,848 & 81.8 \\
50\% & 25,616 & 76.3 \\
75\% & 17,240 & 69.5 \\
\bottomrule
\end{tabular}%
}
\caption{\textbf{Effect of semantic redundancy in the case library.}
The 40,000-case reference configuration matches Table~\ref{tab:library-scale}.
Near-duplicate share is the fraction of cases belonging to clusters containing at least two cases; the cluster count includes singleton clusters.
Macro-F1 is expressed as a percentage.}
\label{tab:semantic-independence}
\end{table}

\subsection{Case-Library Maintenance}
\label{app:library-maintenance}
\textbf{Cold-start initialization.} CaseLaw begins with a cold-start phase in which human moderators assign policy-dependent labels to an initial set of cases. Twenty senior moderators annotate these cases over five rounds. In each round, a random 10\% of annotations undergo double-blind review. A batch is accepted only when annotation agreement exceeds 95\%; otherwise, the affected annotations are revised and reviewed again. This process establishes a reliable seed library for subsequent moderation and incremental expansion.

\textbf{Continual admission from moderation logs.} Once the seed library provides an initial basis for moderation, real cases encountered during online operation are continuously accumulated in a temporary log repository. The repository uses the same record schema as the case library but remains separate from the active retrieval index. At irregular intervals, human moderators audit a sample of logged cases and assess agreement between the recorded labels and their reviewed judgments under the current policy. A batch becomes eligible for admission to the case library when the sampled label agreement reaches at least 95\%. Batches that do not meet this criterion remain in the temporary repository for correction and further review. This staged admission process supports ongoing library growth while maintaining label quality.

\textbf{Threshold calibration before admission.} Each admitted case stores its own similarity threshold $\tau_i$. In our experiments, thresholds for training-set cases are calibrated on the validation set using safe and violating notes. Candidate thresholds are assessed by their effects on violating-note recall and false-positive decisions on safe notes, yielding a case-specific operating point. Cases with more implicit risk cues receive higher thresholds to restrict voting to sufficiently close semantic matches and reduce false positives. During moderation, a retrieved case casts one vote only when its similarity to the query is at least $\tau_i$.

\textbf{Targeted maintenance under policy changes.} When a policy is revised, potentially affected cases are localized through embedding retrieval and keyword search. A batch of representative example notes can be encoded with the frozen embedding model and used as queries to retrieve semantically related cases from the library. Keyword queries over the principal entities or subjects provide a complementary way to locate cases within the scope of the revised rule. Moderators then inspect the retrieved candidates and update the labels and rationales of cases affected by the policy change. Label revisions and affected-case lists are versioned for traceability. Updating labels or rationales preserves existing embeddings, supporting efficient, targeted maintenance.

\section{Cases of Structured Semantic Extraction}
\label{app:qualitative}
\begin{figure}[!t]
\centering\includegraphics[width=\columnwidth]{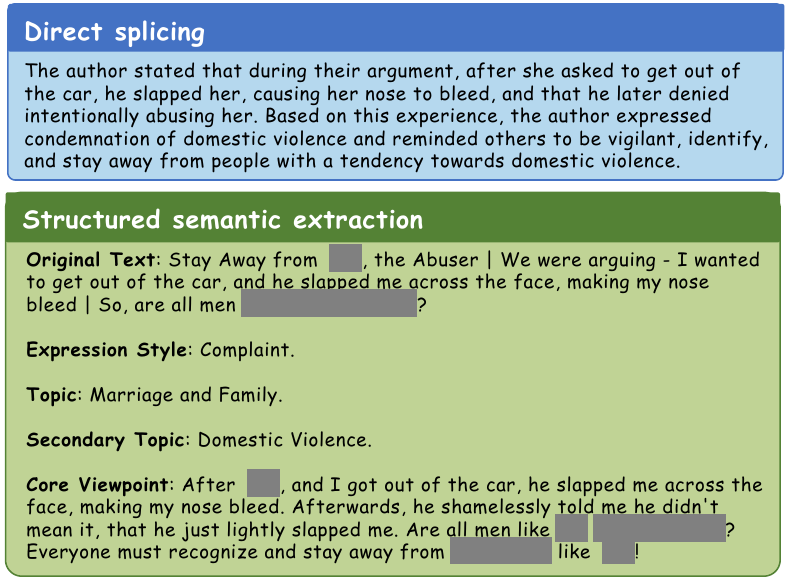}
\caption{Evidence-preserving extraction for an allegation and warning. Identifying and sensitive details are masked.}
\label{fig:sse-case1}
\par\vspace{\floatsep}
\centering\includegraphics[width=\columnwidth]{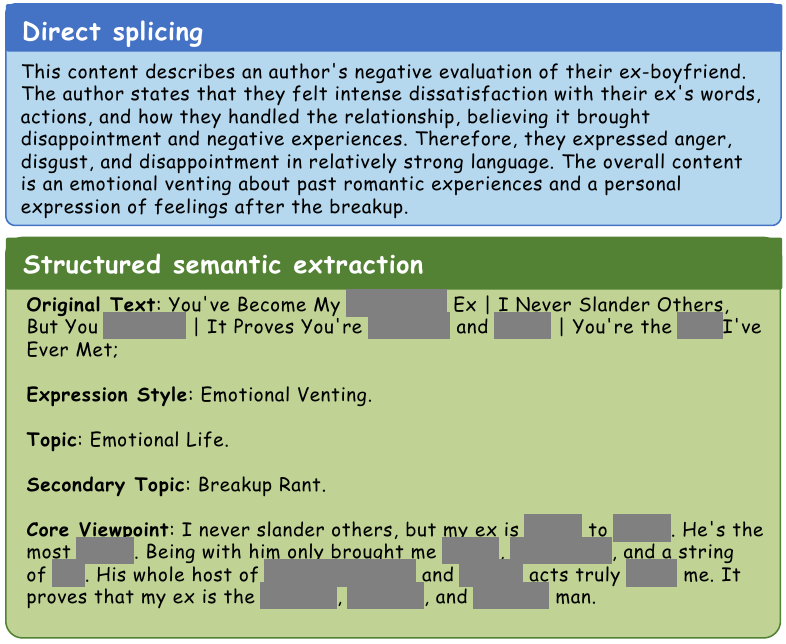}
\caption{Extraction of individual-level emotional expression. Identifying and sensitive details are masked.}
\label{fig:sse-case2}
\end{figure}

Figures~\ref{fig:sse-case1} and~\ref{fig:sse-case2} present two qualitative cases comparing direct splicing with structured semantic extraction from MFO-RedNotes. For ethical and privacy considerations, all personally identifiable information and potentially sensitive details, including privacy-related attributes and violence-related descriptions, are visually masked in the figures. In Case~1, this structure concentrates the representation around the allegation of domestic violence and its warning intent, while in Case~2 it isolates emotional venting in a breakup context from more severe harm-related semantics. By explicitly aligning surface evidence with normalized semantic attributes, structured extraction reduces irrelevant lexical variation and incidental details, leading to more coherent case representations and facilitating more precise retrieval of policy-relevant precedents. This supports CaseLaw's design intuition that structured semantic organization enables a more focused case library for robust moderation under evolving policies.

\section{Metrics}
\label{app:metrics}

\textbf{Moderation performance.} For each class $c$, let $\mathrm{TP}_c$, $\mathrm{FP}_c$, and $\mathrm{FN}_c$ denote one-versus-rest counts over the full evaluation split. Precision, recall, and F1 are
\begin{equation}
\begin{aligned}
P_c &= \frac{\mathrm{TP}_c}{\mathrm{TP}_c+\mathrm{FP}_c},\qquad
R_c = \frac{\mathrm{TP}_c}{\mathrm{TP}_c+\mathrm{FN}_c},\\
F_{1,c} &= \frac{2P_cR_c}{P_c+R_c}.
\end{aligned}
\label{eq:metrics-prf}
\end{equation}
These scores are proportions; percentage reporting multiplies them by 100. For binary evaluation, both ground-truth labels and predictions merge Low-risk, Mid-risk, and High-risk into Unsafe. Abstentions count as false negatives for their true class. Over the evaluated label set $\mathcal Y$, macro-averaging weights classes equally, while worst-label recall measures the least-recovered class:
\begin{equation}
Q_{\mathrm{macro}}=\frac{1}{|\mathcal Y|}\sum_{c\in\mathcal Y}Q_c,
\qquad R_{\min}=\min_{c\in\mathcal Y}R_c,
\label{eq:metrics-macro}
\end{equation}
where $Q\in\{P,R,F_1\}$. Macro-F1 averages class-wise F1 rather than combining macro-precision and macro-recall. For a proportion-valued score $Q$, the absolute gain over baseline set $\mathcal B$ is
\begin{equation}
\Delta Q=100\!\left(Q_{\mathrm{CaseLaw}}-\max_{b\in\mathcal B}Q_b\right),
\label{eq:metrics-gain}
\end{equation}
in percentage points, using the same class, policy, and evaluation setting.

\textbf{False positives and coverage.} Violating-note recall is $R_{\mathrm{Unsafe}}$. With $N_{\mathrm{Safe}}$ safe notes and $N$ total notes, safe-note false-positive rate and automated-decision coverage are
\begin{equation}
\begin{aligned}
\mathrm{FPR}_{\mathrm{Safe}} &= \frac{\mathrm{FP}_{\mathrm{Unsafe}}}{N_{\mathrm{Safe}}},\\
\mathrm{Coverage} &= \frac{1}{N}\sum_{i=1}^{N}\mathbb I\{\hat y_i\ne\bot\},
\end{aligned}
\label{eq:metrics-coverage}
\end{equation}
where $\bot$ denotes abstention or referral for human review.

\textbf{Repeated-run aggregation.} For a fixed method, dataset, and class, let $q^{(r)}$ be precision or recall on the percentage scale in run $r$. Across five runs,
\begin{equation}
\bar q=\frac{1}{5}\sum_{r=1}^{5}q^{(r)},\qquad
s_q^2=\frac{1}{4}\sum_{r=1}^{5}(q^{(r)}-\bar q)^2.
\label{eq:representation-run-statistics}
\end{equation}
The sample variance is expressed in pp$^2$. Aggregation is performed using unrounded run-level scores, with rounding applied only to the reported summaries. A reported variance of 0.00 reflects rounding to two decimal places.

\textbf{Top-five retrieval difference.} Let $\mathcal I_c=\{i:y_i=c\}$, $N_c=|\mathcal I_c|$, and $T_i^{(r)}$ be the unordered set of five highest-ranked case identifiers retrieved before threshold filtering. Top5-diff is
\begin{equation}
D_{5,c}=\frac{100}{N_c}\sum_{i\in\mathcal I_c}
\mathbb I\!\left\{\exists r\in\{2,\ldots,5\}:T_i^{(r)}\ne T_i^{(1)}\right\}.
\label{eq:notecap-top5-diff}
\end{equation}
Each note is counted once if any run changes set membership; rank-only changes do not contribute. The denominator contains ground-truth class-$c$ notes. This diagnostic uses five cases independently of the voting depth $k=16$.

\textbf{Retrieval and label agreement.} Let $R_i^{(r)}$ be the pre-threshold top-$k$ case set. Average pairwise overlap across the ten run pairs is
\begin{equation}
A_{\mathrm{overlap}}=\frac{1}{10N}\sum_{i=1}^{N}
\sum_{1\le r<s\le5}\frac{|R_i^{(r)}\cap R_i^{(s)}|}{k}.
\label{eq:notecap-overlap}
\end{equation}
Strict retrieval-set and label agreement are
\begin{equation}
A_{\mathrm{set}}=\frac{1}{N}\sum_{i=1}^{N}
\mathbb I\{R_i^{(1)}=\cdots=R_i^{(5)}\},
\label{eq:notecap-set}
\end{equation}
\begin{equation}
A_{\mathrm{label}}=\frac{1}{N}\sum_{i=1}^{N}
\mathbb I\{\hat y_i^{(1)}=\cdots=\hat y_i^{(5)}\}.
\label{eq:notecap-label}
\end{equation}
These agreement scores lie in $[0,1]$. Label agreement includes abstention as an outcome; repeated abstention does not imply correct moderation. Retrieved sets may differ while producing identical labels.

\textbf{Annotation agreement.} For a reviewed batch, raw label agreement is $A_{\mathrm{audit}}=n_{\mathrm{match}}/n_{\mathrm{reviewed}}$, where $n_{\mathrm{match}}$ counts matching labels among the reviewed annotations. This measures annotation consistency rather than model accuracy.

\end{document}